\documentclass[11pt]{article}
\usepackage{times}
\usepackage{fullpage}\usepackage[colorlinks=true, allcolors=blue]{hyperref}

\usepackage{array}
\usepackage{graphicx}
\usepackage{multirow}
\usepackage{amsmath,amssymb,amsfonts}
\usepackage{amsthm}
\usepackage{mathrsfs}
\usepackage{bbm}
\usepackage[title]{appendix}
\usepackage{xcolor}
\usepackage{textcomp}
\usepackage{manyfoot}
\usepackage{booktabs}
\usepackage{algorithmicx}
\usepackage{algpseudocode}
\usepackage{listings}
\usepackage{xspace}
\usepackage{caption}
\usepackage{mathtools}
\usepackage[capitalise]{cleveref}
\usepackage{subcaption}
\usepackage{authblk}

\DeclareCaptionLabelSeparator{bar}{\space\textbf{\textbar}\space}
\DeclareCaptionFont{captionfont}{\fontsize{11pt}{11pt}\selectfont}
\newcommand{\decisive}{decisive patterns\xspace}
\newcommand{\sensitive}{sensitive patterns\xspace}
\newcommand{\Sensitive}{Sensitive patterns\xspace}
\newcommand{\Decisive}{Decisive patterns\xspace}
\newcommand{\SEnsitive}{Sensitive Patterns\xspace}
\newcommand{\DEcisive}{Decisive Patterns\xspace}

\newcommand{\acts}{$\operatorname{ActsTrack}$\xspace}
\newcommand{\taumu}{$\operatorname{Tau3Mu}$\xspace}
\newcommand{\plb}{$\operatorname{PLBind}$\xspace}
\newcommand{\syn}{$\operatorname{SynMol}$\xspace}

\newcommand{\fidelityp}{Fidelity+ AUC\xspace}
\newcommand{\fidelityn}{Fidelity- AUC\xspace}
\newcommand{\fidelity}{Fidelity AUC\xspace}
\newcommand{\labelfidelity}{Decisive-Induced Fidelity AUC\xspace}

\newcommand{\first}[1]{\mathbf{#1}^\dagger}
\newcommand{\second}[1]{\mathbf{#1}}
\newcommand{\third}[1]{\underline{#1}}

\begin{document}

\title{Towards Understanding Sensitive and Decisive Patterns in Explainable AI: A Case Study of Model Interpretation in Geometric Deep Learning}


\author[1]{Jiajun Zhu\footnote{Zhu was an intern at Georgia Institute of Technology when doing this project.}}
\author[2]{Siqi Miao}
\author[3]{Rex Ying}
\author[2]{Pan Li\footnote{Correspondence: panli@gatech.edu}}


\affil[1]{Zhejiang University, Hangzhou, China}
\affil[2]{Georgia Institute of Technology, Atlanta, US}
\affil[3]{Yale University, New Haven, US}



\date{}
\maketitle
\begin{abstract}
The interpretability of machine learning models has gained increasing attention, particularly in scientific domains where high precision and accountability are crucial. This research focuses on distinguishing between two critical data patterns—sensitive patterns (model-related) and decisive patterns (task-related)—which are commonly used as model interpretations but often lead to confusion. Specifically, this study compares the effectiveness of two main streams of interpretation methods: post-hoc methods and self-interpretable methods, in detecting these patterns. Recently, geometric deep learning (GDL) has shown superior predictive performance in various scientific applications, creating an urgent need for principled interpretation methods. Therefore, we conduct our study using several representative GDL applications as case studies. We evaluate thirteen interpretation methods applied to three major GDL backbone models, using four scientific datasets to assess how well these methods identify sensitive and decisive patterns. Our findings indicate that post-hoc methods tend to provide interpretations better aligned with sensitive patterns, whereas certain self-interpretable methods exhibit strong and stable performance in detecting decisive patterns. Additionally, our study offers valuable insights into improving the reliability of these interpretation methods. For example, ensembling post-hoc interpretations from multiple models trained on the same task can effectively uncover the task's decisive patterns.

\end{abstract}
\section{Introduction}

Machine learning (ML) methods can make accurate predictions with a data-driven approach, exhibiting significant promise for various scientific applications~\cite{butler2018molecular, carleo2019physical, zhong2021machine, bergen2019machine}. 
Among these methods, geometric deep learning (GDL) has emerged as a revolutionary approach, especially in the domains where data naturally form point clouds, such as particle clouds in high energy physics (HEP)~\cite{qu2020jet, ju2021performance}, proteins in biochemistry~\cite{gainza2020deciphering, stark2022equibind}, and molecules in material science~\cite{liao2023equiformerv2, zhou2023uni}.
GDL models have shown remarkable predictive performance in many such applications 
since they excel at learning representations from point cloud data
by preserving geometric equivariance and incorporating domain-specific inductive bias~\cite{schutt2017schnet,jing2020learning,bogatskiy2020lorentz,sanchez2020learning}. 
However, the black-box nature 
hinders the understanding of these models' decision-making processes. 
This highlights the urgent need for interpretable 
GDL models,  especially for those employed in scientific applications, where both high precision and accountability are paramount~\cite{doshi2017towards}.

\begin{figure}[t]
    \centering
    \caption{Overview of GDL model interpretation and its evaluation:
    Interpretation in geometric deep learning (GDL) tasks involves identifying a subset of points $C_s$ from the input point cloud $C$. \Decisive are a subset of points that inherently dictate the labels of the point cloud, specified by the learning task, and their identification accuracy is measured by the alignment between $C_s$ and the true \decisive (Interpretation ROC-AUC). \Sensitive, on the other hand, are the subset of most influential points affecting the model’s predictions, as specified by the model itself. The evaluation of the model’s sensitivity involves assessing the changes of its predictions when $C_s$ is either added to or removed from the input (Fidelity AUC).}
    \label{fig:overview}
    \includegraphics[
    trim={0.9cm 0.6cm 1.8cm 0.4cm},clip, 
    width=\textwidth]{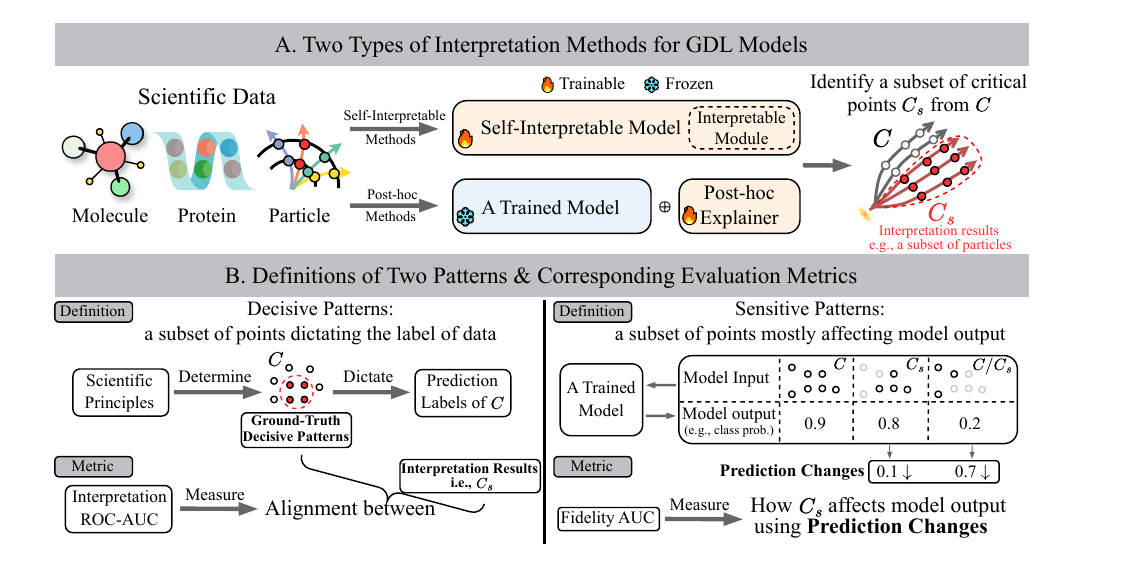}
\end{figure}

In terms of model interpretation, two patterns in the data are relevant yet often confused by researchers, which we name as \emph{\sensitive} and \emph{\decisive} (see Fig.~\ref{fig:overview}B). Conceptually, \sensitive are
those whose presence or absence greatly influences the model's predictions. \Sensitive may vary among different models that tackle the same learning task. 
\Decisive, on the other hand, are \emph{intrinsic to the learning task} and 
determine the labels of the prediction task, regardless of the specific model being used. 
Despite their conceptual distinction, existing studies rarely distinguishes between them when evaluating different interpretation approaches.  
Most previous works focus solely on detecting \sensitive~\cite{puiutta2020explainable,madsen2022post,danilevsky2020survey,jacovi2020towards,linardatos2020explainable}, while several other works hypothesize the alignment of the two patterns and use them exchangeably 
~\cite{yuan2021subgraphx,lucic2022cf,yuan2022explainability,vu2020pgm,miao2022interpretable}. This confusion risks misunderstanding evaluation outcomes. For instance, it might involve employing label-relevant data patterns (i.e., \decisive) to assess the quality of extracted \sensitive for a given ML model or leveraging the \sensitive a model detects to gain insights into the underlying learning task~\cite{luo2020parameterized}. Hence, a systematic exploration of the connections and disparities between these two patterns, particularly concerning the capabilities of current model interpretation methods to detect them, is imperative.





There are mainly two categories of methods designed to provide ML models with interpretability~\cite{arrieta2020explainable, rudin2019stop, yuan2022explainability} (see Fig.~\ref{fig:overview}A). The first category, known as \emph{post-hoc} methods, operates on already-trained models and aims to interpret their predictive behaviors. Post-hoc methods may conceptually extract \sensitive, as the extracted patterns are specific to those already-trained models. The second category comprises \emph{self-interpretable} methods, which often integrate interpretable modules into model architectures and optimize these modules during the model training. These interpretable modules are likely to better uncover \decisive by sharing the goal of accurately predicting the labels of the task. 
Nonetheless, the merits and drawbacks of these two categories of methods have sparked ongoing debates and controversies~\cite{rudin2019stop, laugel2019dangers}. In particular, 
there is a lack of systematic investigation and comparison on the ability of these methods to detect the two types of data patterns respectively.
Most previous studies limit their scope to only one category, either post-hoc methods~\cite{yuan2022explainability, amara2022graphframex, sanchez2020evaluating, longa2022explaining, chen54generative, chen2024d4explainer, laugel2019dangers,adebayo2021post,slack2021reliable, bui2024explaining} or self-interpretable methods~\cite{serrano2019attention,jain2019attention,mohankumar2020towards,bai2021attentions,wang2023pursuit}.
Some recent surveys~\cite{li2022survey, wu2022survey, kakkad2023survey, zhang2022trustworthy} have overviewed both post-hoc and self-interpretable approaches, however, these surveys primarily focus on establishing taxonomies of different interpretability methods and fail to further compare and evaluate the two categories of approaches. As a result, the pros and cons of these approaches in detecting the two data patterns remain unresolved questions.



With the growing need for interpretability in scientific fields, this study uses GDL models that are notably prevalent in these areas~\cite{qu2020jet, ju2021performance, gainza2020deciphering, stark2022equibind, liao2023equiformerv2, zhou2023uni} as the testbeds to address the above questions. 
Considering there have been 
no studies on GDL model interpretation except \cite{miao2022lri} to the best of our knowledge, 
our study also contributes by extending many current methods originally proposed for other models to GDL models and evaluating them systematically in a highly modular software platform published together with this study\footnote{\url{https://github.com/Graph-COM/xgdl}}. Given that geometric point cloud data can be represented as graphs by connecting close points in space, interpretability methods designed for the models that encode graph data, e.g., graph neural networks (GNNs)~\cite{ying2019gnnexplainer, luo2020parameterized, yuan2021subgraphx}, could be extended for comparison in our study. Specifically, we adapt 11 established interpretation methods for GNNs to the GDL setting by incorporating geometric features and application-specific principles such as symmetries of the underlying physical systems. 
In total, we benchmark 13 interpretation methods with 3 important GDL backbone models to evaluate their abilities to extract \sensitive and \decisive. The model interpretation pipeline and its evaluation regarding each type of patterns, are illustrated in Figure~\ref{fig:overview}.

Our study is based on 4 scientific datasets from applications in high-energy physics (HEP)~\cite{ai2022common, cms2020search} and biochemistry~\cite{mccloskey2019using, wang2005pdbbind}. In these fields, GDL methods have proven extremely effective but urgently need reliable interpretability. These datasets are either collected from actual experiments or from reliable simulations that are extensively used in their corresponding domains. Contrary to previous studies that often use datasets lacking ground-truth labels for \decisive or generated by simple rules (e.g., motif-based)~\cite{ying2019gnnexplainer, luo2020parameterized, chen2024tempme, yuan2021subgraphx, vu2020pgm}, the datasets we employed are more realistic and are annotated with \decisive according to the underlying scientific principles. This setup allows us to assess the methods' capability to extract \decisive and to measure the alignment between \decisive and \sensitive.



Our extensive evaluation yields solid evidence that the interpretations given by post-hoc methods generally align well with \sensitive but not \decisive. 
In contrast, the interpretations given by some self-interpretable methods, such as LRI-induced methods~\cite{miao2022lri}, align well with \decisive. 
In addition, we also observe that some post-hoc methods may face instability issues, i.e., the same method may demonstrate inconsistent performance across different datasets. The performance of self-interpretable methods can be more stable but method-dependent: some self-interpretable methods can effectively identify both \decisive and \sensitive, whereas others may fail to discern either. 

Besides the above high-level observations, our studies elaborate more insights by answering the following three questions:
\textbf{Q1:} Given that the interpretations yielded by post-hoc methods do not align well with the \decisive, what strategies can enhance the alignment to potentially enable post-hoc methods to detect \decisive for the learning tasks? 
\textbf{Q2:} Do the \sensitive of models trained based on self-interpretable methods align well with the \decisive of the task? In other words, are self-interpretable models inherently sensitive to \decisive? \textbf{Q3:} Whether and how the degree of alignment between the \sensitive and the \decisive is influenced by the quality (e.g., prediction accuracy) of models to be interpreted?
Specifically, the insights based on our extensive evaluation for 
the above questions, 
along with some broader implications and significance, are summarized as follows: 
\begin{enumerate}
    \item
    We observe that the interpretations given by post-hoc methods vary greatly among different models even when models were trained in the same setting and achieved high prediction accuracy but just used different random seeds. This indicates a fundamental limitation of post-hoc methods to detect \decisive for the learning tasks.  Nonetheless, we address the problem by ensembling the interpretations yielded for multiple trained models. 
    The ensembled interpretations align much better with \decisive, which enables post-hoc methods to more reliably uncover \decisive for the learning tasks.

    \item The \sensitive of some self-interpretable models may align well with the \decisive, which indicates that such models can be robust to non-decisive artifacts in the datasets and are mostly sensitive to the \decisive.
    This implies that well-performed self-interpretable methods may produce more reliable and interpretable models and may be a more favorable choice compared to post-hoc methods.

    \item Models with higher predictive accuracy tend to have better alignment between their \sensitive and \decisive for the learning tasks, suggesting that as predictive performance improves, a model's predictive behavior becomes increasingly influenced by the \decisive. Consequently, robust label predictive performance is a foundational prerequisite if both the \sensitive of the model and the \decisive for the learning task are desired.
    
\end{enumerate}


\section{Results}

\begin{figure}[tbp]
    \centering
    \caption{An overview of the interpretation methods benchmarked. Our evaluation considers two major categories: post-hoc and self-interpretable methods. Within the post-hoc methods, four sub-categories are further segmented based on the techniques employed. Similarly, the self-interpretable methods are organized into three distinct categories, each also differentiated by the used techniques.
    }
    \label{fig:baselines}
    \includegraphics[
    trim={0cm 3.5cm 0cm 3cm},clip, 
    width=\textwidth]{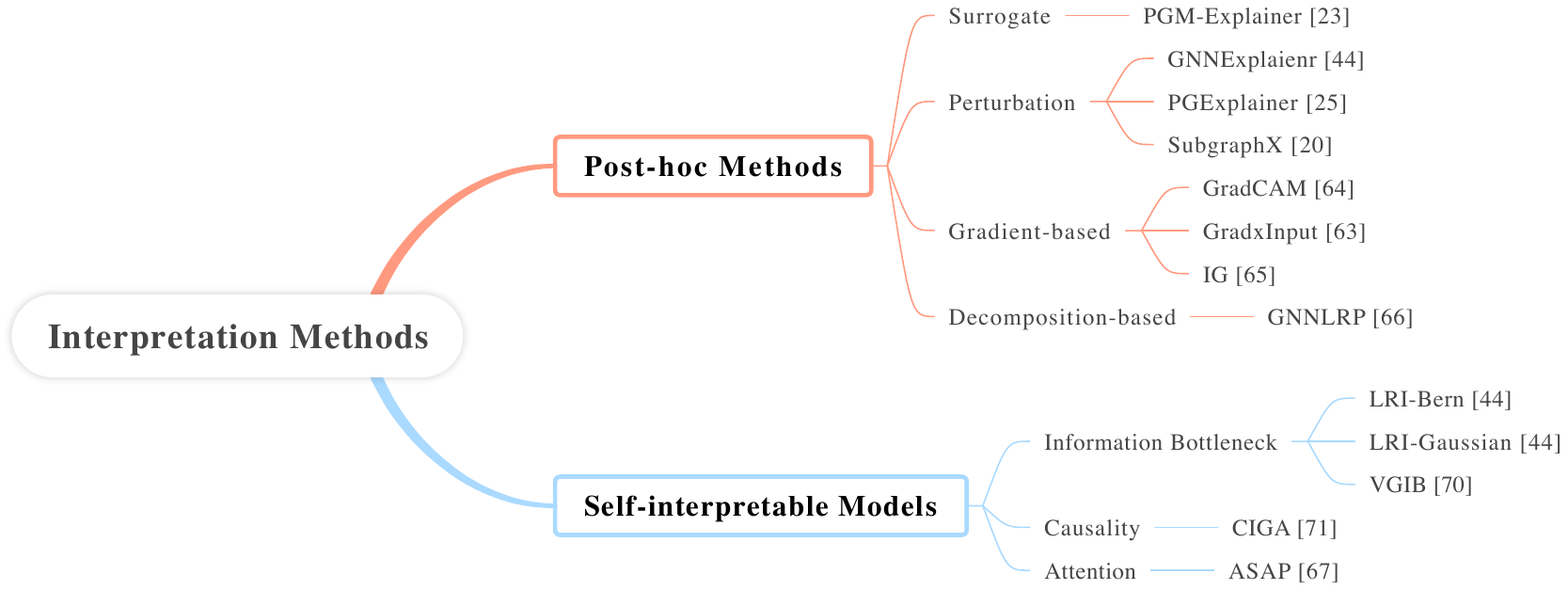}
\end{figure}

\subsection{Evaluation Framework}

\textbf{Model Interpretation Task.} 
We focus on GDL tasks where each data sample is characterized as a point cloud, denoted by $ C = (\mathcal{V}, \mathbf{X}, \mathbf{p})$. Here, $ \mathcal{V} = \{ v_1, v_2, \ldots, v_n \} $ represents a collection of $ n $ points, $ \mathbf{X} \in \mathbb{R}^{n \times d} $ comprises $ d $-dimensional features for each point, and $ \mathbf{p} \in \mathbb{R}^{n \times k} $ specifies 2D or 3D spatial coordinates for these points, depending on the specific application.
Each sample $C \in \mathcal{C}$  is associated with a class label $Y\in \mathcal{Y}$, and GDL models \( f_{\theta}(\cdot): \mathcal{C} \rightarrow \mathcal{Y} \) are trained to make a prediction $\hat{Y}$ for each test data instance.
Model interpretation is presented as a meaningful subset of points $ C_s = (\mathcal{V}_s, \mathbf{X}_s, \mathbf{p}_s)$ from $C$ output by an interpretation method. 
In practice, $C_s$ is typically obtained via the following process: An interpretation method will assign a list of importance scores $\mathcal{W} = \{ w_1, w_2, \ldots, w_n \}$ where $w_i\in \mathbb{R}$ is for every individual point $v_i\in C$, and output the top-ranked points given a selection ratio $\rho$ as the subset $C_s$, e.g., for $\rho=0.2$, the top-ranked 20\% critical points in $C$ will form $C_s$.



\noindent\textbf{Decisive Patterns.} Decisive patterns depend on the learning tasks, which inherently determine the class label $Y$ according to the specific scientific principles and are model-irrelevant. Specifically, each point in $C$ also comes with a binary label $z_i$ denoting if it is part of the \decisive, collected by $\mathcal{I} = \{ z_1, z_2, \ldots, z_n \}$. 
For instance, in the \taumu dataset (shown later), each sample $C$ has the class label $Y$ indicating the occurrence of decay $\tau \rightarrow \mu\mu\mu$ in $C$, and those points representing the $\mu$'s from this decay are labeled as $z_i=1$. 
Note that $\mathcal{I}$ is not used during model training but serves exclusively for evaluation by testing the alignment between the results output by interpretation algorithms and $\mathcal{I}$. 

\noindent \textbf{\SEnsitive.} \Sensitive for point cloud data are defined as the most influential subsets of points on the model predictions. 
Specifically, the influence of any subset is quantified by assessing whether the model prediction changes upon the removal of the subset from the initial data, or whether it remains unchanged when only the subset is provided as input. Mathematically, these two criteria are respectively expressed as $\mathop{\arg\max}_{C_s} \left| f_{\theta}(C) - f_{\theta}(C \backslash C_s) \right|$ and $\mathop{\arg\min}_{C_s} \left| f_{\theta}(C) - f_{\theta}(C_s) \right| $, where $f_{\theta}$ represents the model and $C_s$ is typically constrained with a budget on its size. In this study, we consider a combination of these two criteria $\mathop{\arg\max}_{C_s}\left| f_{\theta}(C) - f_{\theta}(C \backslash C_s) \right|-\left| f_{\theta}(C) - f_{\theta}(C_s) \right| $ as the definition of \sensitive. Note that unlike \decisive that are inherently determined by scientific principles and can be labeled accordingly, \sensitive are defined as model-specific and are the patterns captured by each model during training, which can vary across different models. 

\noindent\textbf{Scientific Datasets.}
We briefly introduce 4 GDL datasets~\cite{miao2022lri} derived for real-world scientific applications employed in our experiments as below, the details of which can be found in Sec.~\ref{subsec:datasets}.
\begin{itemize}
    \item \textbf{\acts} is to reconstruct the properties of charged particles  using position measurements from tracking detectors. This process is essential for numerous downstream analyses in HEP, such as identifying particle types and reconstructing collision events~\cite{thomson2013modern, ai2022common}. Unlike traditional track reconstruction, the task here involves predicting the presence of $\mu$ tracks from a $z \rightarrow \mu\mu$ decay and the detector hits left by $\mu$ corresponds to decisive patterns.
    
    \item \taumu is another HEP dataset, focusing on the detection of the rare and challenging signature of the $\tau \rightarrow \mu\mu\mu$ decay, which is highly suppressed in the Standard Model of particle physics~\cite{oerter2006theory, Blackstone_2020}. Therefore, the detection of such decays is a strong indicator of potential new physics~\cite{calibbi2018charged, atlas2020search}. The decisive patterns in this dataset are the detector hits from $\mu$.
    
    \item \syn aims at molecular property prediction, hypothesizing that molecules containing specific functional groups could bind to a target protein. Accurate prediction of molecular properties can significantly accelerate drug design and substance discovery efforts~\cite{faber2017prediction}. The decisive patterns in this dataset are the atoms in two functional groups: carbonyl and unbranched alkane.
    
    \item \plb is used to predict protein-ligand binding affinities based on the 3D structures of proteins and ligands. It is a crucial step because a high affinity is one of the major drug selecting criteria~\cite{wang2017improving, karimi2019deepaffinity}. The decisive patterns here are the amino acids located in the binding site of the test protein.
\end{itemize}

\vspace{1mm}
\noindent\textbf{Benchmarked Methods.}
We select three GDL backbone models for interpretation evaluation, namely  EGNN~\cite{satorras2021n}, DGCNN~\cite{wang2019dynamic}, and PointTrans ~\cite{zhao2021point}, as they are widely employed in scientific applications~\cite{qu2020jet,atz2021geometric,gagliardi2022shrec}.
As depicted in Fig.~\ref{fig:baselines}, our benchmark includes a total of 13 interpretability methods, which represent a broad spectrum of techniques to provide an inclusive evaluation. We provide a brief introduction of them as follows according to the taxonomy in~\cite{yuan2022explainability}. 
The detailed description of each method and the strategy used to adapt the graph-based methods to GDL tasks can be found in Sec.~\ref{sec:method}.

\emph{Post-hoc methods} generate interpretation results for an already-trained model $f_\theta$ by assigning an importance score for each point.
Among the four main categories of post-hoc methods, gradient-based and decomposition-based approaches directly utilize the properties of the already-trained model, such as gradients, while perturbation and surrogate methods build an extra learnable explainer $g_\phi$ to assign point importance. Specifically, gradients-based methods, including GradxInput~\cite{shrikumar2017learning}, GradCAM~\cite{chattopadhay2018grad} and Integrated Gradients (IG) ~\cite{sundararajan2017axiomatic}, compute the gradients with respect to the inputs or the learned (intermediate) point embeddings to identify important points. Decomposition-based methods, e.g., GNNLRP~\cite{schnake2021gnnlrp}, devise score decomposition rules to distribute the prediction scores layer by layer in a back-propagation manner to the input space for identifying points that impact prediction scores the most.
Perturbation methods, including GNNExplainer~\cite{ying2019gnnexplainer}, PGExplainer~\cite{luo2020parameterized} and SubgprahX~\cite{yuan2021subgraphx}, 
propose different approaches to perturb inputs and train an explainer $g_\phi$ to select important input patterns according to the output variations of $f_\theta$.
Surrogate methods, such as PGM-Explainer~\cite{vu2020pgm}, train another interpretable surrogate model (e.g., probabilistic graphical model) to locally approximate the predictions of the original model and use the trained surrogate model to understand the decision-making process of the original model. Notably, all these post-hoc methods will not change the original $f_{\theta}$ in any way.

\emph{Self-interpretable methods}, on the other hand, design interpretable modules $g_{\phi}$ and integrate such modules into existing backbone models $f_{\theta}$. The combined models $f_\theta \circ g_\phi$ are then trained from scratch and are self-interpretable due to the integrated interpretable modules.
The common self-interpretable methods include the following three categories. Attention-based methods, such as ASAP~\cite{ranjan2020asap}, regard attention distributions as an interpretation, where the part of inputs with higher attention weights are believed to have a greater influence on the model's decision-making process. The other two categories of methods are based on principles such as the information bottleneck (IB)~\cite{tishby2000information} and causality analysis~\cite{pearl2009causality}. The IB-based methods, including LRI-induced methods~\cite{miao2022lri} and VGIB~\cite{yu2022vgib}, design interpretable modules $g_{\phi}$ to restrict information flow and encourage the model $f_\theta \circ g_\phi$ to extract input patterns that are with minimal sufficient information for the task.
Causality-based methods, such as CIGA~\cite{chen2022ciga}, assume causal relationships within the data remain unchanged across different environments and aim to extract such invariant data patterns using $g_\phi$.

For post-hoc methods, we begin by pre-training GDL backbone models (i.e., DGCNN, EGNN, and Point Transformer), each with 10 distinct random seeds per dataset, using the cross-entropy loss, and apply all post-hoc methods to each of the already-trained models. 
The same seeds are used again if the post-hoc methods include any parameters to be optimized in a data-driven way. As self-interpretable methods integrate their interpretable modules into the chosen GDL backbones, self-interpretable models are trained from scratch, each with 10 random seeds as well, using the objective functions specified by each method. 

As far as we know, the only methods currently tailored specifically for GDL models are LRI-induced methods~\cite{miao2022lri}, i.e., LRI-Bern and LRI-Gaussian, and other methods are adapted by us to the GDL setting. Note that PGM-Explainer~\cite{vu2020pgm} and SubgraphX~\cite{yuan2021subgraphx} are only evaluated on two of the datasets, i.e., \syn and \acts, due to their extreme inefficiency. For example, PGM-Explainer/SubgraphX requires more than 20/72 hours to train one seed on \taumu dataset using an NVIDIA RTX A6000 GPU.
\vspace{1mm}

\vspace{1mm}

\vspace{1mm}
\noindent\textbf{Evaluation Metrics.} 
We employ three widely-used metrics to assess how well a method can extract \sensitive and \decisive.

To quantify each method's effectiveness in identifying \decisive, we utilize \emph{Interpretation ROC-AUC}~\cite{luo2020parameterized} for  the three datasets \syn, \acts and \taumu, and \emph{Precision@20}~\cite{miao2022lri} for the \plb dataset. 
Interpretation ROC-AUC is calculated by comparing the 
\decisive $\mathcal{I}$ with the importance scores $\mathcal{W}$. 
Precision@20 is gauged by the ratio of points in the decisive patterns 
among the top 20 ranked points. Higher Interpretation ROC-AUC or Precision@20 indicates the detected subset of points $C_s$ align better with the \decisive.


To evaluate the ability of each method to identify \sensitive, we use \emph{Fidelity AUC}~\cite{yuan2022explainability} for four datasets.
This involves calculating Fidelity+ and Fidelity- given a subset \( C_s \) derived from the importance scores \( \mathcal{W} \) provided by an interpretation method, 
which correspond to the two criteria used in defining \sensitive, respectively.
Fidelity+ is calculated by taking the average value over the test dataset of the expression \(\mathbbm{1}(f_{\theta}(C) = Y) - \mathbbm{1}(f_{\theta}(C \setminus C_s) = Y)\) for each test data instance $C$. A higher Fidelity+ indicates more sensitivity of $f_{\theta}$ to the pattern $C_s$ . Fidelity- is determined by the formula 
\(\mathbbm{1}(f_{\theta}(C) = Y) - \mathbbm{1}(f_{\theta}(C_s) = Y)\), with lower averages reflecting more sensitivity. The overall Fidelity score is defined as the arithmetic mean of Fidelity+ and \( 1 - \) Fidelity-. To comprehensively evaluate performance, we vary 
the sizes of $C_s$ and compute Fidelity AUC as the area under the curve of Fidelity (versus the size of $C_s$). 
A higher Fidelity AUC suggests that the model's interpretations are more closely aligned with \sensitive.
For a more detailed description of the calculation of Fidelity AUC, we refer readers to Sec.~\ref{sec:exp}.

\subsection{Benchmarking Interpretability Performance}
\begin{table*}[t]
\centering
\caption{Fidelity AUC and Interpretation ROC-AUC or Precision@20 performance of the 13 methods. The $\textbf{Bold}$ and $\underline{\text{Underline}}$ highlight the first and second best results within each category of methods. The $\textbf{Bold}^\dagger$ highlights the best results across all methods in terms of Interpretation ROC-AUC or Precision@20. The results are reported as mean $\pm$ std.
}
\label{tab:main_res}

\resizebox{\textwidth}{!}{%
\begin{tabular}{@{}lcccccccccccc@{}}%
\toprule%
\multicolumn{1}{c}{\multirow{3}{*}{Method}} & \multicolumn{6}{c}{\syn} & \multicolumn{6}{c}{\acts} \\
\cmidrule(lr){2-7} \cmidrule(lr){8-13}
& \multicolumn{3}{c}{\fidelity} & \multicolumn{3}{c}{Interpretation ROC AUC} & \multicolumn{3}{c}{\fidelity} & \multicolumn{3}{c}{Interpretation ROC AUC} \\
\cmidrule(lr){2-4} \cmidrule(lr){5-7} \cmidrule(lr){8-10} \cmidrule(lr){11-13}
& EGNN & DGCNN & PointTrans & EGNN & DGCNN & PointTrans & EGNN & DGCNN & PointTrans & EGNN & DGCNN & PointTrans \\
\midrule%
GNNLRP & $\third{76.28}\pm1.85$ & $\second{61.69}\pm9.54$ & $49.93\pm2.66$ & $\third{81.75}\pm4.01$ & $\second{84.61}\pm3.81$ & $50.38\pm1.68$ & $\third{93.41}\pm1.96$ & $\second{90.39}\pm2.01$ & $50.48\pm11.32$ & $\second{86.01}\pm2.31$ & $\third{86.40}\pm5.08$ & $50.20\pm1.93$ \\
GradCAM & $57.62\pm4.52$ & $51.14\pm2.27$ & $66.39\pm2.62$ & $57.82\pm4.42$ & $78.89\pm3.84$ & $84.10\pm3.66$ & $\second{93.55}\pm3.07$ & $80.23\pm3.77$ & $\third{89.75}\pm4.07$ & $\third{69.38}\pm2.72$ & $75.25\pm3.67$ & $\third{77.32}\pm2.83$ \\
GradxInput & $71.11\pm5.04$ & $52.02\pm2.38$ & $68.31\pm1.82$ & $76.03\pm4.82$ & $71.39\pm5.89$ & $78.03\pm1.52$ & $79.19\pm1.89$ & $72.58\pm3.43$ & $79.80\pm3.08$ & $68.74\pm1.84$ & $65.17\pm1.56$ & $64.78\pm1.90$ \\
IG & $72.98\pm7.12$ & $50.66\pm1.10$ & $\third{72.71}\pm2.38$ & $78.59\pm7.83$ & $64.31\pm9.14$ & $\third{84.23}\pm1.83$ & $79.22\pm1.86$ & $72.49\pm3.41$ & $79.68\pm3.17$ & $68.78\pm1.82$ & $65.27\pm1.47$ & $64.80\pm1.88$ \\
GNNExplainer & $58.55\pm12.91$ & $50.03\pm0.54$ & $27.60\pm2.80$ & $58.94\pm15.89$ & $51.03\pm5.58$ & $26.28\pm2.70$ & $50.71\pm16.03$ & $69.10\pm5.47$ & $87.00\pm2.74$ & $51.77\pm4.41$ & $64.34\pm4.05$ & $71.38\pm2.61$ \\
PGExplainer & $67.62\pm16.36$ & $51.06\pm2.02$ & $66.17\pm2.17$ & $77.92\pm22.04$ & $49.56\pm39.82$ & $\second{87.41}\pm2.66$ & $28.18\pm40.3$ & $\third{89.83}\pm5.59$ & $\second{95.24}\pm2.61$ & $33.54\pm23.17$ & $\second{92.63}\pm1.57$ & $\second{88.39}\pm3.13$ \\
PGM-Explainer & $60.40\pm2.55$ & $50.45\pm0.41$ & $53.83\pm1.29$ & $64.59\pm0.87$ & $51.45\pm1.57$ & $58.99\pm0.75$ & $80.37\pm3.21$ & $54.44\pm2.24$ & $56.17\pm2.32$ & $62.89\pm1.13$ & $55.06\pm1.29$ & $55.00\pm1.02$ \\
SubgraphX & $\second{88.06}\pm1.28$ & $\third{59.13}\pm7.86$ & $\second{80.20}\pm1.22$ & $\second{86.7}\pm1.78$ & $\third{68.26}\pm4.33$ & $77.82\pm1.00$ & $92.00\pm2.91$ & $83.68\pm3.40$ & $86.57\pm1.42$ & $62.93\pm2.86$ & $60.08\pm0.05$ & $62.78\pm0.94$ \\
\midrule
ASAP & $59.25\pm6.70$ & $65.68\pm6.18$ & $51.32\pm4.66$ & $64.20\pm11.69$ & $79.55\pm6.82$ & $57.10\pm8.54$ & $\second{92.2}\pm4.93$ & $88.49\pm4.97$ & $64.63\pm14.77$ & $\third{81.03}\pm4.25$ & $89.00\pm2.22$ & $64.69\pm13.35$ \\
CIGA & $55.38\pm11.2$ & $51.12\pm3.09$ & $47.29\pm14.05$ & $62.19\pm21.23$ & $61.90\pm22.34$ & $51.21\pm21.69$ & $36.22\pm37.22$ & $48.07\pm37.99$ & $36.51\pm37.68$ & $43.98\pm21.15$ & $47.31\pm31.47$ & $40.53\pm24.11$ \\
LRI-Bern & $\third{80.97}\pm3.82$ & $\third{74.74}\pm3.76$ & $\third{70.41}\pm1.94$ & $\third{92.04}\pm3.00$ & $\third{94.20}\pm4.53$ & $\third{90.46}\pm1.21$ & $87.63\pm2.19$ & $\third{90.52}\pm1.84$ & $\second{92.31}\pm1.22$ & $80.97\pm2.07$ & $\third{90.74}\pm1.72$ & $\third{86.84}\pm1.85$ \\
LRI-Gaussian & $\second{82.97}\pm3.26$ & $\second{77.26}\pm2.67$ & $\second{75.12}\pm1.72$ & $\first{97.13}\pm0.79$ & $\first{98.23}\pm1.00$ & $\first{93.06}\pm1.19$ & $\third{90.93}\pm3.85$ & $\second{91.12}\pm1.55$ & $\third{90.17}\pm3.19$ & $\first{92.93}\pm1.58$ & $\first{94.18}\pm0.88$ & $\first{91.85}\pm1.15$ \\
VGIB & $79.02\pm3.05$ & $53.13\pm2.73$ & $56.82\pm1.93$ & $88.19\pm3.23$ & $93.88\pm6.51$ & $72.03\pm2.96$ & $71.09\pm20.48$ & $78.20\pm8.11$ & $74.39\pm18.87$ & $60.27\pm11.06$ & $90.30\pm2.12$ & $72.31\pm15.31$ \\
\bottomrule%
\end{tabular}%
}

\vspace{1mm}
\resizebox{\textwidth}{!}{%
\begin{tabular}{@{}lcccccccccccc@{}}%
\toprule%
\multicolumn{1}{c}{\multirow{3}{*}{Method}} & \multicolumn{6}{c}{\taumu} & \multicolumn{6}{c}{\plb} \\%
\cmidrule(lr){2-7} \cmidrule(lr){8-13}%
 & \multicolumn{3}{c}{\fidelity} & \multicolumn{3}{c}{Interpretation ROC AUC} & \multicolumn{3}{c}{\fidelity} & \multicolumn{3}{c}{Precision@20} \\%
\cmidrule(lr){2-4} \cmidrule(lr){5-7} \cmidrule(lr){8-10} \cmidrule(lr){11-13}%
 & EGNN & DGCNN & PointTrans & EGNN & DGCNN & PointTrans & EGNN & DGCNN & PointTrans & EGNN & DGCNN & PointTrans \\%
\midrule%
GNNLRP & $\third{61.96}\pm1.61$ & $\second{63.68}\pm1.52$ & $49.13\pm0.45$ & $76.00\pm0.82$ & $\second{73.15}\pm1.9$ & $50.00\pm0.21$ & $82.47\pm6.01$ & $\second{75.41}\pm7.77$ & $50.04\pm4.44$ & $\second{67.17}\pm9.03$ & $\first{63.34}\pm5.33$ & $45.41\pm2.94$ \\%
GradCAM & $61.03\pm2.65$ & $60.66\pm1.60$ & $\third{62.34}\pm0.82$ & $74.49\pm3.69$ & $\third{68.48}\pm3.58$ & $\first{80.72}\pm0.80$ & $79.27\pm6.56$ & $\third{73.33}\pm5.87$ & $70.03\pm10.55$ & $57.89\pm6.57$ & $\third{60.65}\pm5.20$ & $57.48\pm6.29$ \\%
GradxInput & $61.14\pm1.98$ & $\third{63.28}\pm1.30$ & $61.78\pm0.79$ & $\second{77.50}\pm2.36$ & $68.25\pm0.27$ & $68.69\pm0.42$ & $\third{83.21}\pm7.28$ & $65.30\pm9.43$ & $74.97\pm5.56$ & $60.60\pm3.67$ & $55.91\pm7.01$ & $\second{59.10}\pm5.22$ \\%
IG & $61.45\pm2.36$ & $60.37\pm0.80$ & $60.66\pm0.79$ & $\second{77.50}\pm2.62$ & $65.39\pm0.26$ & $68.35\pm0.43$ & $\second{85.97}\pm5.84$ & $65.84\pm13.50$ & $76.90\pm5.22$ & $\third{60.68}\pm4.31$ & $53.15\pm6.82$ & $\third{57.58}\pm5.36$ \\%
GNNExplainer & $61.94\pm1.22$ & $48.17\pm2.85$ & $36.45\pm1.23$ & $71.98\pm2.38$ & $52.51\pm3.29$ & $30.47\pm0.77$ & $52.4\pm10.25$ & $44.56\pm11.09$ & $\second{77.96}\pm5.57$ & $42.27\pm3.69$ & $44.15\pm3.14$ & $57.57\pm4.01$ \\%
PGExplainer & $\second{62.09}\pm1.66$ & $48.92\pm11.88$ & $\second{62.74}\pm0.87$ & $\third{76.10}\pm2.32$ & $52.24\pm19.53$ & $\third{78.60}\pm0.55$ & $59.8\pm29.83$ & $55.87\pm28.57$ & $\third{77.10}\pm7.49$ & $56.36\pm9.97$ & $47.82\pm10.03$ & $55.93\pm6.53$ \\%
\midrule
ASAP & $56.27\pm1.45$ & $52.58\pm3.80$ & $50.10\pm0.24$ & $66.63\pm1.67$ & $69.54\pm0.61$ & $52.15\pm3.77$ & $50.14\pm0.30$ & $49.87\pm0.22$ & $50.19\pm0.34$ & $45.05\pm0.16$ & $45.10\pm0.00$ & $45.10\pm0.00$ \\%
CIGA & $51.00\pm17.51$ & $43.32\pm11.84$ & $48.74\pm10.2$ & $54.07\pm23.31$ & $43.95\pm19.53$ & $49.72\pm17.67$ & $49.26\pm1.91$ & $49.24\pm9.62$ & $50.79\pm1.94$ & $45.55\pm7.68$ & $41.75\pm4.40$ & $48.74\pm5.54$ \\%
LRI-Bern & $\third{63.07}\pm1.79$ & $\second{65.52}\pm1.67$ & $\third{62.83}\pm0.99$ & $\third{77.51}\pm2.79$ & $\third{78.23}\pm1.26$ & $\third{78.02}\pm0.82$ & $\third{50.94}\pm3.36$ & $\third{56.99}\pm2.77$ & $51.83\pm3.93$ & $\third{72.14}\pm4.89$ & $\third{61.75}\pm7.72$ & $\first{68.86}\pm7.81$ \\%
LRI-Gaussian & $\second{63.60}\pm1.44$ & $\third{64.62}\pm0.95$ & $\second{63.76}\pm1.28$ & $\first{80.48}\pm0.49$ & $\first{81.41}\pm0.63$ & $\second{79.88}\pm0.46$ & $\second{54.97}\pm4.87$ & $50.26\pm4.65$ & $\third{55.40}\pm8.55$ & $\first{74.40}\pm0.64$ & $\second{63.16}\pm5.25$ & $\third{57.70}\pm5.70$ \\%
VGIB & $59.86\pm3.91$ & $58.92\pm5.86$ & $51.37\pm13.53$ & $73.88\pm5.56$ & $72.70\pm11.80$ & $54.16\pm26.22$ & $47.14\pm9.64$ & $\second{66.63}\pm4.05$ & $\second{75.24}\pm6.76$ & $46.16\pm6.91$ & $54.66\pm6.34$ & $56.31\pm6.00$ \\%
\bottomrule%
\end{tabular}%
}

\end{table*}

In this subsection, we benchmark 8 post-hoc and 5 self-interpretable methods to evaluate their effectiveness in extracting \sensitive and \decisive. It is important to note that since \sensitive are model-specific, comparing the extraction capabilities of \sensitive between post-hoc and self-interpretable methods is not appropriate (all post-hoc methods work on the same already-trained models, but self-interpretable methods would train new models from scratch using their proposed objectives). Below we briefly describe the results presented in Table.~\ref{tab:main_res}. 

\subsubsection{Benchmarking Post-Hoc Methods}

Regarding the performance of extracting \sensitive of the already-trained models, SubgraphX outperforms most other post-hoc methods across backbone models and datasets. Its success is likely due to its unique, albeit computationally intensive, approach that employs Monte Carlo tree search to identify important points. 
However, its computational complexity limits its applicability to larger datasets like \taumu and \plb. 
Following closely is GNNLRP, which demonstrates strong performance on various datasets, but its performance declines when applied to Point Transformer. We speculate this drop stems from the manually crafted propagation rules needed by GNNLRP, which may conflict with the architecture of Point Transformer.
PGExplainer represents a more intricate case. While it can achieve the third-best Fidelity AUC when its results are stable, i.e., with low variances, it may fail on models trained with certain random seeds, especially when applied to EGNN.
Among gradient-based approaches, which yield consistent results across all settings likely because they do not involve a separate learning phase, only GradCAM stands out as competitive.
On the downside, GNNExplainer and PGM-Explainer underperform in our benchmark, revealing their limitations in effectively extracting \sensitive in the GDL tasks considered. 

Regarding the performance of extracting \decisive of the learning tasks, GNNLRP achieves leading Interpretation ROC-AUC scores when it is not paired with Point Transformer. Surprisingly, this time GradCAM performs rather competitively, surpassing SubgraphX in most cases, while other gradient-based methods still perform subpar. As for PGExplainer, again, it provides unstable results with high variances, even though it shows great performance in a few cases. GNNExplainer and PGM-Explainer, similarly, do not seem to work well in our experiments.

\subsubsection{Benchmarking Self-Interpretable Models}
Although self-interpretable methods are not designed to detect \sensitive for a given model, it is still interesting to see whether the models trained by self-interpretable methods are sensitive to their extracted interpretation patterns. Notably, LRI-Bern and LRI-Gaussian achieve relatively high Fidelity AUC scores. As for the remaining models, VGIB overall performs the third best but suffers from high variances on some datasets, ASAP occasionally exhibits high Fidelity AUC scores but generally lags behind, while CIGA appears ill-suited when adapted to the GDL even with significant parameter tuning.

As for the extraction of \decisive, LRI-Bern and LRI-Gaussian consistently deliver superior performance in all settings, significantly outperforming other methods, including post-hoc ones. VGIB and ASAP follow LRI-induced methods in performance, yet ASAP demonstrates high variances across different settings.  

\subsubsection{Comparing Post-Hoc and Self-Interpretable Methods}

To summarize, when comparing the ability to identify \decisive, although post-hoc methods, notably SubgraphX and GNNLRP, may often offer decent results, top-performed self-interpretable methods, e.g., LRI-Gaussian, significantly outperform all post-hoc methods, suggesting using the output interpretations of self-interpretable methods when one cares more about the \decisive for the learning tasks.
Furthermore, the generally poor Interpretation ROC-AUC performance of post-hoc methods, in contrast to their relatively high Fidelity AUC, indicates that post-hoc interpretations may not align well with the \decisive, and we will further investigate this issue in Sec.~\ref{sec:align}.

Note that one cannot directly compare post-hoc and self-interpretable methods regarding their capabilities of detecting \sensitive, as the models to be interpreted are revised when one applies self-interpretable methods. 
Nonetheless, we can still see a trend that self-interpretable methods achieving better Interpretation ROC-AUC (the metric for detecting \decisive) typically obtain better Fidelity AUC (the metric for detecting \sensitive). Moreover, as the achieved Fidelity AUC scores of some self-interpretable methods are generally comparable with those yielded by post-hoc methods, the models trained based on self-interpretable methods  are also sensitive to the interpretations these methods output. 

\subsection{Relationship of Post-Hoc Extracted Interpretations and \DEcisive}\label{sec:align}

Besides our benchmark, in this section, we study the question (Q1): Given that the interpretations given by post-hoc methods do not align well with the \decisive (i.e., post-hoc methods tend to exhibit poor performance regarding Interpretation ROC-AUC despite having high \fidelity), what strategies can enhance the alignment to potentially enable post-hoc methods to detect \decisive for the learning tasks?




\begin{figure}[t]
  \centering
  \includegraphics[width=\textwidth]{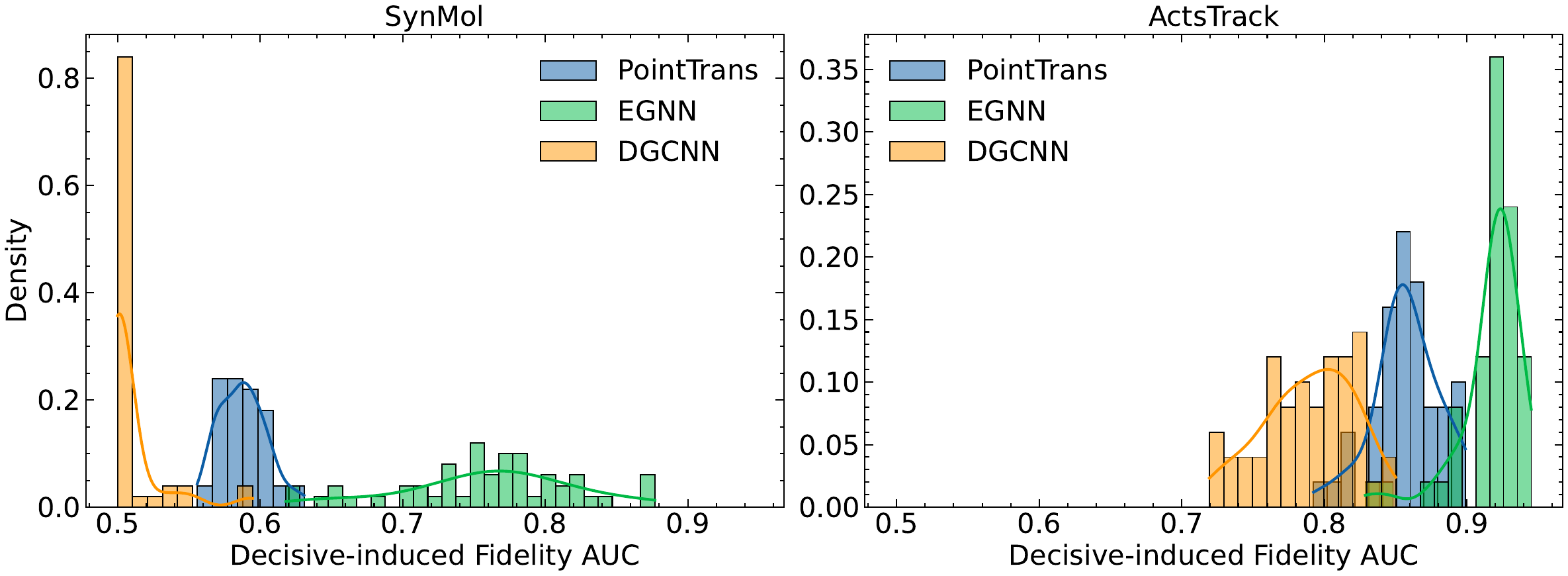}
  \caption{\labelfidelity of various models for the three backbone models. 50 models were trained for each backbone per dataset.}
  \label{fig:notalign}
\end{figure}

\subsubsection{Investigating the General Misalignment Between \SEnsitive and \DEcisive}

We hypothesize the misalignment between post-hoc interpretations and \decisive is essentially caused by the general misalignment between the \sensitive of a model and the \decisive of the learning tasks, 
as post-hoc methods are more designed to detect those model-specific \sensitive.  
Based on this hypothesis, we propose to check the \fidelity when directly inputting the labeled ground-truth \decisive for each sample as identified important points, which we term as \labelfidelity.
This is built on the assumption that the model should be sensitive to decisive patterns if the two patterns are well-aligned. We assess the \labelfidelity across 50 models for each backbone using the \syn and \acts datasets, and we visualize the distribution of \labelfidelity. 



As shown in Figure~\ref{fig:notalign}, the \labelfidelity scores are generally low across diverse datasets and backbones and are even significantly lower than the \fidelity scores of the corresponding post-hoc methods (Supplementary Table~\ref{tab:notalign}). Note that here all the models are well trained and some of them may even have accuracy as high as 98\% (Supplementary Table~\ref{tab:clf_perf}). 
These observations imply a fundamental misalignment between the two patterns for the GDL models. Moreover, \labelfidelity scores also exhibit substantial standard deviations, suggesting that models trained with different random seeds have significantly varying levels of sensitivity to \decisive.
This highlights the need to distinguish between the objectives of extracting \sensitive and \decisive in practical applications. Specifically, if the goal is to identify what patterns a model is sensitive to, post-hoc methods such as GNNLRP prove effective. However, if the goal is to derive knowledge from the data by extracting \decisive, one should be conservative when applying post-hoc methods since they may produce interpretations that misalign with \decisive.


\subsubsection{The Ensemble Strategy to Improve the Alignment}

\begin{table*}
\centering
\caption{Performance of extracting \decisive using post-hoc methods with the ensemble strategy. Numbers in the parentheses indicate the improvement upon the Interpretation ROC-AUC reported in Table~\ref{tab:main_res}.}
\label{tab:ens_auc}

\resizebox{\linewidth}{!}{%
\begin{tabular}{@{}lcccccccccccc@{}}
\toprule
\multirow{2}{*}{Method} & \multicolumn{3}{c}{SynMol} & \multicolumn{3}{c}{ActsTrack} & \multicolumn{3}{c}{Tau3Mu} & \multicolumn{3}{c}{PLbind} \\
\cmidrule(lr){2-4} \cmidrule(lr){5-7} \cmidrule(lr){8-10} \cmidrule(lr){11-13}
 & EGNN & DGCNN & PointTrans & EGNN & DGCNN & PointTrans & EGNN & DGCNN & PointTrans & EGNN & DGCNN & PointTrans \\
\midrule
GNNExplainer & $70.98 (12.04)$ & $45.49 (-5.54)$ & $39.92 (13.64)$ & $62.47 (10.70)$ & $75.73 (11.39)$ & $80.08 (8.70)$ & $77.52 (5.54)$ & $57.22 (4.71)$ & $41.64 (11.17)$ & $49.10 (6.83)$ & $46.30 (2.15)$ & $61.50 (3.93)$ \\
GNNLRP & $84.42 (2.67)$ & $87.26 (2.65)$ & $53.62 (3.24)$ & $89.97 (3.96)$ & $89.76 (3.36)$ & $53.25 (3.05)$ & $78.87 (2.87)$ & $75.99 (2.84)$ & $49.75 (-0.25)$ & $74.40 (7.23)$ & $62.20 (-1.14)$ & $50.60 (5.19)$ \\
GradCAM & $63.38 (5.56)$ & $82.93 (4.04)$ & $90.12 (6.02)$ & $75.05 (5.67)$ & $80.19 (4.94)$ & $86.15 (8.83)$ & $80.50 (6.01)$ & $71.95 (3.47)$ & $83.06 (2.34)$ & $62.10 (4.21)$ & $61.80 (1.15)$ & $58.70 (1.22)$ \\
GradxInput & $82.42 (6.39)$ & $78.41 (7.02)$ & $82.67 (4.64)$ & $69.61 (0.87)$ & $65.43 (0.26)$ & $65.08 (0.30)$ & $80.14 (2.64)$ & $68.20 (-0.05)$ & $69.35 (0.66)$ & $62.40 (1.80)$ & $57.90 (1.99)$ & $63.70 (4.60)$ \\
IG & $91.87 (13.28)$ & $69.78 (5.47)$ & $87.48 (3.25)$ & $69.65 (0.87)$ & $65.49 (0.22)$ & $65.11 (0.31)$ & $79.29 (1.79)$ & $65.41 (0.02)$ & $68.70 (0.35)$ & $63.70 (3.02)$ & $60.60 (7.45)$ & $59.60 (2.02)$ \\
PGExplainer & $96.20 (18.28)$ & $94.69 (45.13)$ & $90.51 (3.10)$ & $63.59 (30.05)$ & $95.01 (2.38)$ & $91.71 (3.32)$ & $78.60 (2.50)$ & $71.90 (19.66)$ & $80.01 (1.41)$ & $64.90 (8.54)$ & $62.50 (14.68)$ & $61.30 (5.37)$ \\
PGM-Explainer & $68.83 (4.24)$ & $53.31 (1.86)$ & $63.94 (4.95)$ & $70.26 (7.37)$ & $58.52 (3.46)$ & $61.58 (6.58)$ & - & - & - & - & - & - \\
SubgraphX & $92.32 (5.62)$ & $76.84 (8.58)$ & $82.38 (4.56)$ & $64.60 (1.67)$ & $60.21 (0.13)$ & $63.93 (1.15)$ & - & - & - & - & - & - \\
\bottomrule
\end{tabular}%
}

\end{table*}

The above misalignment disqualifies using post-hoc interpretations as the \decisive of the learning tasks. However, an interesting question is that if the significant variation in the \sensitive of the models gets removed, can we safely treat post-hoc interpretations as approximation of the \decisive? 

Therefore, we propose employing an ensemble of the post-hoc interpretations for multiple already-trained models as a strategy to enhance the extraction of \decisive. In our experiments, we apply each post-hoc method on 10 models trained with different seeds, resulting in 10 importance scores for each point in the point cloud $C$. Then, we utilize a \emph{weighted average aggregation} to yield a final score for each point. The weight is determined by the fidelity of each post-hoc explainer, calculated as $\mathop{\min} \{0, \text{\fidelity} - 50\}$ and then normalized. Note that this weight neither relies on the classification label $Y$ nor the labels of ground-truth \decisive $\mathcal{I}$. The quality of this emsemble strategy is evaluated in Interpretation ROC-AUC or Precision@20, as shown in Table~\ref{tab:ens_auc}.

On average, the ensemble method significantly improves the identification of \decisive based on post-hoc interpretations by 12.97\%, 9.42\%, 7.02\%, and 8.43\% on \syn, \acts, \taumu, and \plb datasets, respectively. The most significant boost is observed for the \syn dataset, which could potentially arise from the numerous spurious correlations (i.e., correlations between the irrelevant input environment
features and the labels) within this dataset~\cite{miao2022lri}. Spurious correlations are likely to be captured by the models,  subsequently being extracted as the interpretations by post-hoc methods yet essentially irrelevant to \decisive. 
The ensemble strategy helps with filtering out these irreverent non-decisive patterns. 
We claim that ensembling the post-hoc interpretations across multiple already-trained models is necessary. To see this, we also evaluate an ensemble of multiple post-hoc interpretations generated based on multiple random seeds but for the same already-trained model (see Supplementary Table \ref{tab:more_ens}), which yields much worse performance than that in Table~\ref{tab:ens_auc}. 

\subsection{Are \SEnsitive of Self-Interpretable Models  Aligned Well with \DEcisive?}

\begin{table*}[t]
\centering
\caption{Interpretation of ROC-AUC values derived from post-hoc methods applied to self-interpretable models trained via LRI-induced methods. In each setting, scores are averaged across three model backbones. An \underline{underline} indicates that the score is lower than that of the corresponding ERM model. For comprehensive results with individual backbones, refer to the Supplementary Table~\ref{tab:more_post_inherent}.}
\label{tab:exp_comparison}
\resizebox{\linewidth}{!}{
\begin{tabular}{@{}lcccccc@{}}
\toprule
\multirow{2}{*}{Interpretation Method} & \multicolumn{3}{c}{\syn} & \multicolumn{3}{c}{\acts} \\
\cmidrule(lr){2-4} \cmidrule(lr){5-7}
& ERM Model & LRI-Bern Model & LRI-Gaussian Model & ERM Model & LRI-Bern Model & LRI-Gaussian Model \\
\midrule
GNNLRP       & $72.25 \pm 9.50$ & $\third{71.89} \pm 9.53$ & $74.08 \pm 4.22$ & $74.20 \pm 9.32$ & $75.81 \pm 8.41$ & $76.41 \pm 9.13$ \\
GradCAM      & $73.60 \pm 11.92$ & $\third{73.48} \pm 14.41$ & $75.15 \pm 15.62$ & $73.98 \pm 9.22$ & $77.32 \pm 8.82$ & $77.50 \pm 12.24$ \\
GradxInput   & $75.15 \pm 12.23$ & $83.07 \pm 14.08$ & $83.41 \pm 12.66$ & $66.23 \pm 5.30$ & $66.52 \pm 5.19$ & $74.48 \pm 6.65$ \\
IG           & $75.71 \pm 18.80$ & $82.10 \pm 17.05$ & $86.14 \pm 9.81$ & $66.28 \pm 5.17$ & $66.61 \pm 5.22$ & $74.50 \pm 6.58$ \\
GNNExplainer & $45.42 \pm 24.17$ & $47.16 \pm 0.29$ & $49.84 \pm 0.03$ & $62.50 \pm 11.07$ & $63.41 \pm 0.10$ & $\third{49.97} \pm 0.03$ \\
PGExplainer  & $71.63 \pm 64.52$ & $77.70 \pm 0.96$ & $\third{42.12} \pm 0.45$ & $71.52 \pm 27.87$ & $87.42 \pm 0.18$ & $85.59 \pm 0.18$ \\
Self         & $-$ & $92.23 \pm 8.74$ & $96.14 \pm 2.98$ & $-$ & $86.18 \pm 5.64$ & $92.99 \pm 3.61$ \\
\bottomrule
\end{tabular}
}
\end{table*}


Given that the interpretations given by LRI-induced methods also demonstrate high \fidelity (Table~\ref{tab:main_res}), it might indicate that the models trained by LRI-induced methods are already sensitive to the \decisive. In other words, the \sensitive and \decisive are potentially well aligned for these models. 
To verify this conjecture, 
we apply post-hoc methods to the models trained by LRI-induced methods 
and evaluate the Interpretation ROC-AUC by comparing the obtained post-hoc interpretations with the decisive patterns of these tasks. The Interpretation ROC-AUC of the model that has the same architecture but goes through standard training pipelines is used as a baseline. 


As illustrated in Table~\ref{tab:exp_comparison}, using \emph{any} post-hoc method, the post-hoc interpretations of LRI-induced models consistently demonstrate better alignment with the \decisive, compared to the post-hoc interpretations of the same model architectures but trained via standard empirical risk minimization (ERM). 
Notably, these interpretations often show a significant improvement, with gains reaching up to 15.9\%. These observations support the claim that LRI-induced models are inherently sensitive to decisive patterns.

\begin{figure}[htbp]
    \centering
    \begin{subfigure}[t]{\textwidth} %
        \centering
        \includegraphics[width=1.0\textwidth]{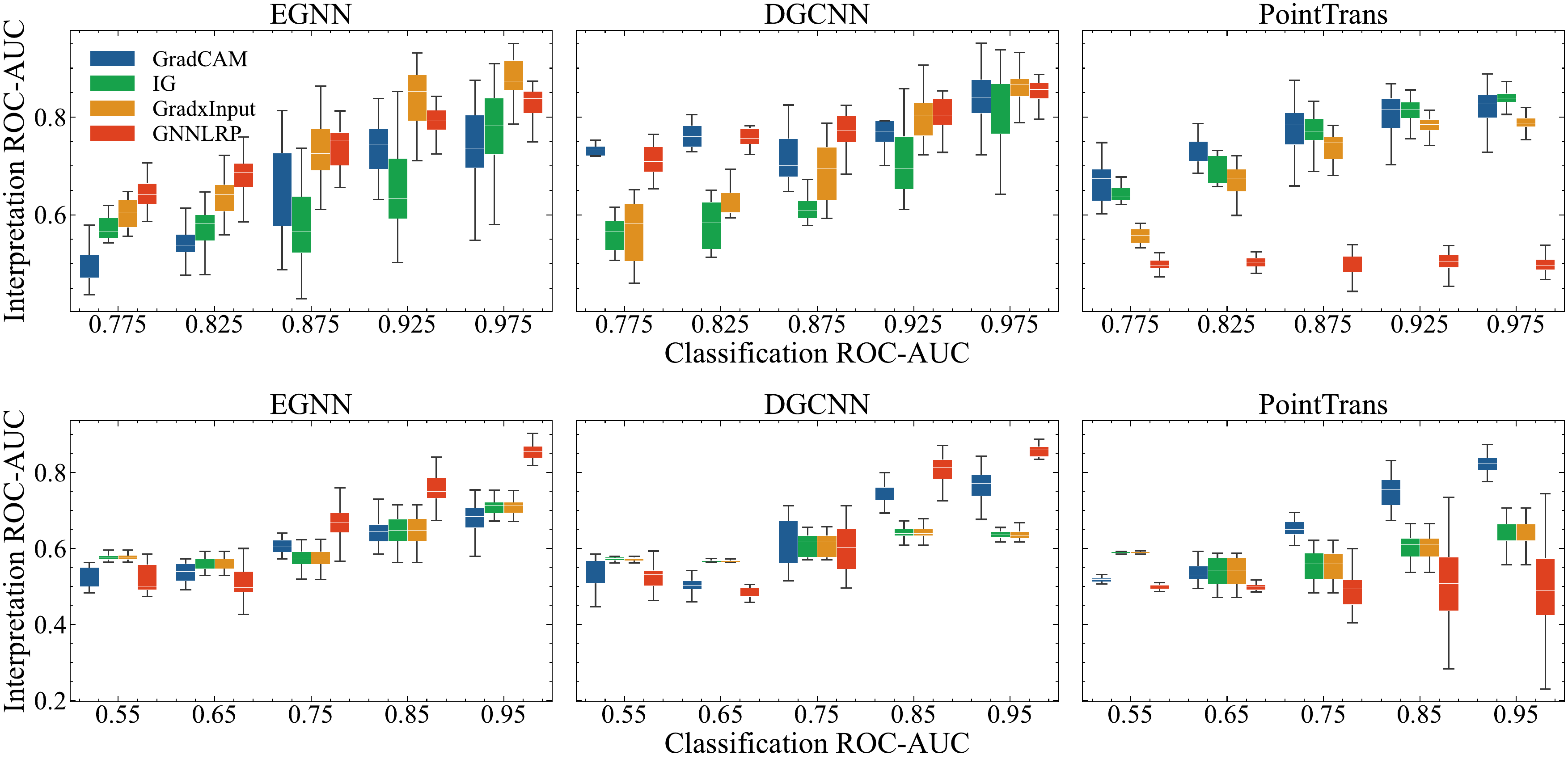}
        \caption{Classification ROC-AUC v.s. Interpretation ROC-AUC for four post-hoc methods. The first row corresponds to results from \syn, followed by \acts in the subsequent row.}
        \label{fig:clf_exp_post}
        \vspace{2mm}
    \end{subfigure}
    \begin{subfigure}[t]{\textwidth} %
        \centering
        \includegraphics[width=1.0\textwidth]{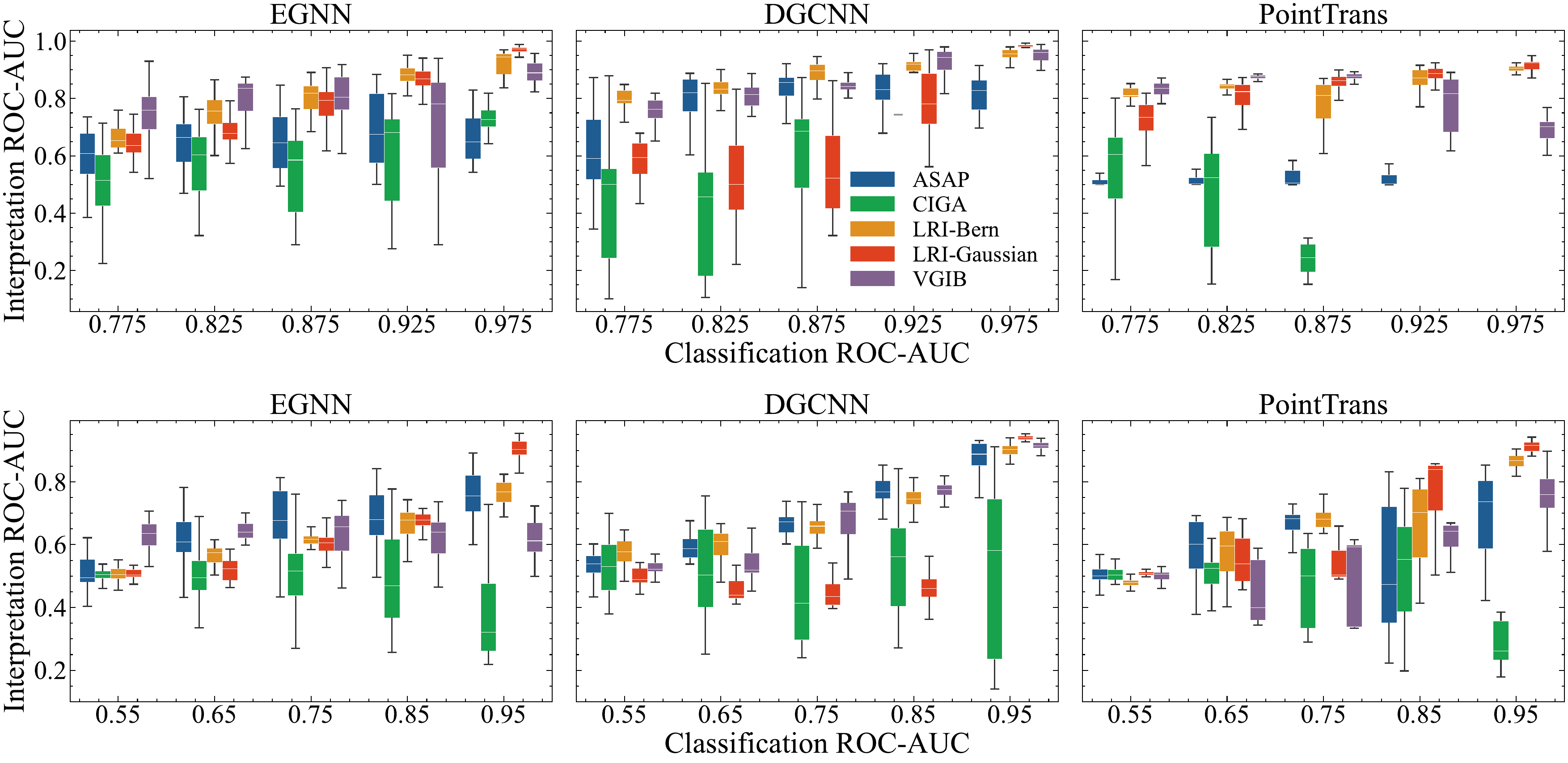}
        \caption{Classification ROC-AUC v.s. Interpretation ROC-AUC for five self-interpretable methods. The first row corresponds to results from \syn, followed by \acts in the subsequent row.}
        \label{fig:clf_exp_inhe}
        \vspace{2mm}
    \end{subfigure}
    \caption{Classification ROC-AUC v.s. Interpretation ROC-AUC for both post-hoc and self-interpretable methods on \syn and \acts datasets.}
    \label{fig:clf_exp}
\end{figure}

\subsection{Model Prediction Accuracy Indicates the Alignment Between the Two Patterns}

Here, we explore how prediction accuracy impacts the interpretation results of the models (Q3).
Specifically, we train 150 models for each dataset and backbone with various training recipes, resulting in models having a wide range of classification accuracy. Due to the large number of trained models for this study, we run the most efficient four post-hoc methods and summarize the  Interpretation ROC-AUC results in Fig.~\ref{fig:clf_exp_post}. Note that for each dataset, we divide the models according to their classification accuracy into five intervals and draw four box plots for the models within each interval based on Interpretation ROC-AUCs given by the four post-hoc methods. 
In parallel, we also trained the models based on five self-interpretable methods and similarly presented the results in Fig.~\ref{fig:clf_exp_inhe}.


As shown in Fig.~\ref{fig:clf_exp_post}, when the model's classification performance improves, the Interpretation ROC-AUC performance of post-hoc methods tends to improve in a similarly linear manner. Similarly, the increasing trend depicted in Fig.~\ref{fig:clf_exp_inhe} is evident.
This suggests that a model would indeed be more sensitive to \decisive when achieving better prediction accuracy. This makes sense because when the model captures the \decisive for the learning task, it tends to generalize better. Therefore, high model prediction accuracy can be viewed as a good indicator if one would like to detect the \decisive of the learning task by analyzing the \sensitive of the model.

\section{Discussion}
\noindent\textbf{Main Conclusions.} 
This work has systematically investigated two important categories of model interpretation approaches (post-hoc interpretation v.s. self-interpretation) for GDL models, regarding their capabilities of detecting two types of data patterns (\sensitive v.s. \decisive) that were often confused by previous model interpretation studies. 
\Sensitive are model-specific and post-hoc methods present reasonable performance when detecting them. Our evaluation shows that SubgraphX among all post-hoc methods achieves the best \sensitive' extraction. \decisive are task-specific and independent of the learning models. Self-interpretable methods can produce better and more stable interpretation results when detecting \decisive. Among self-interpretable methods, LRI-Gaussian often achieves the best performance. 

Our investigation reveals the fundamental misalignment between post-hoc interpretations and \decisive, which is mainly caused by variations in the \sensitive of the pre-trained models. We observe that high model prediction accuracy often serves as a strong indicator of alignment between the \sensitive of the pre-trained models and the \decisive of the task. 
We also propose an ensemble method that combines post-hoc interpretations of multiple pre-trained models to improve the alignment between the post-hoc  interpretations and \decisive. 
Furthermore, our investigation finds that models trained by some self-interpretable methods may inherently be more sensitive to \decisive compared to models trained in standard ERM. 
\vspace{1mm}
\noindent\textbf{Significance for ML Researchers and Domain Scientists.}
Our results contribute to the advancement of studying interpretability in ML methods and its applications in scientific domains. The observed misalignment between the two patterns underscores the necessity for subsequent ML researchers to clearly establish their objectives of identifying \sensitive or \decisive before applying or devising any interpretability methods. Our observations also indicate that it can be inappropriate to directly compare methods designed for different objectives. For example, it can be unfair to compare the ability of post-hoc methods to uncover \decisive with self-interpretable methods, suggesting the need for different evaluation frameworks for different methods.

For domain scientists, our study highlights post-hoc methods to be better-suited for validating the reliability of already-trained models, i.e., checking whether a model's predictive behavior is mainly sensitive to patterns related to established scientific principles, while if the goal is to uncover (unknown) insights and knowledge from the data, self-interpretable methods designed to extract \decisive can be a better choice.
Moreover, for a task where no models can achieve high prediction accuracy, trying to uncover decisive patterns using interpretability methods may be futile.
Due to the relative infancy of GDL as a field, extensive analysis of GDL methods for science remains scarce, with even fewer studies on interpretable GDL for scientific purposes.
Therefore, we have established a solid foundation for researchers in this promising direction and have paved the way for the development of a trustworthy GDL pipeline for scientific applications.

\vspace{1mm}
\noindent\textbf{Limitations.}
Our work has several potential limitations in the context of ML for science. First, we have not taken into consideration task-specific backbone models which are popular alternative models in scientific ML and whose interpretability may be of significant value to domain scientists.
Second, in addition to the performance metrics used in this study, there exist several other metrics meaningful for evaluating the identification of the two types of patterns, such as the consistency across different backbones~\cite{sanchez2020evaluating}, the fairness in multilabel classification~\cite{dai2022fairness}, and the stability for perturbed counterpart~\cite{agarwal2023evaluating}, etc. These are beyond our focus and may further complement this study in the future.
Third, due to the limited number of interpretability methods specially designed to directly work on the geometric features in GDL, our benchmark only examines the selection of a subset of points as the model interpretation, while missing elaborating how the distribution of this subset of points in the space may present the model  interpretation. 
However, GDL models may capture more fine-grained geometric patterns because of the nature of GDL tasks, therefore, the examination of geometric coordinates may provide more scientific insights from a different perspective~\cite{miao2022lri}. 




\section{Methods}
In this section, we will provide more details on the scientific applications and the datasets, along with some implementation details of our experiments. We will also describe the various interpretation methods applied in our study. 


\subsection{More Details on Interpretation Methods}\label{sec:method}


\subsubsection{Post-Hoc Methods}
Given a pre-trained model, post-hoc methods generate model interpretations without modifying the model's learned weights.
They are primarily categorized into five groups. From each category, we select one of the representative methods and extend it to GDL models.

\vspace{1mm}
\noindent\textbf{Gradients-Based Methods}~\cite{shrikumar2017learning, chattopadhay2018grad, sundararajan2017axiomatic, knyazev2019understanding, baldassarre2019explainability} compute gradients with respect to the input or intermediate activations to explain trained models.
\emph{GradxInput}~\cite{shrikumar2017learning} calculates the element-wise product between the input features and the gradient with respect to them to measure the importance of different input patterns. 
\emph{GradCAM}~\cite{chattopadhay2018grad} extends GradxInput by using intermediate activations, multiplying the gradients of activations with the activations themselves to derive importance scores. \emph{Integrated Gradients} (IG)~\cite{sundararajan2017axiomatic} 
assigns importance values to each input feature by integrating the gradients with respect to the input across a path from a non-informative input to the actual input.
In GDL models, gradients are computed with respect to the input point coordinates for IG and GradXInput. For GradCAM, gradients are derived from intermediate activations.

\vspace{1mm}
\noindent\textbf{Decomposition-Based Methods}~\cite{montavon2019layer, schnake2021gnnlrp, xiong2022efficient, pope2019explainability} build score decomposition rules to distribute the prediction scores layer by layer in a back-propagation manner to the input space to identify points that contribute the most to the prediction scores.
Inspired by layer-wise relevance propagation (LRP) algorithm~\cite{montavon2019layer}, \emph{GNNLRP}~\cite{schnake2021gnnlrp} studies the importance of different walks in the graph (i.e. sequences of edges) to interpret GNNs. The importance of each edge is determined by considering all graph walks that contain it.
To adapt the method from GNNs to GDL, we extend the framework to evaluate the significance of points within the point cloud. This is achieved by aggregating the importance scores derived from the walks traversed.

\vspace{1mm}
\noindent\textbf{Perturbation Methods}~\cite{ying2019gnnexplainer, luo2020parameterized, yuan2021subgraphx, schlichtkrull2020interpreting, anonymous2021hard,bui2024explaining} generate perturbation masks to select important
input by optimizing the output variations of the trained models with respect
to different input perturbations.
\emph{GNNExplainer}~\cite{ying2019gnnexplainer} employs soft masks, learned through mask optimization for individual input graphs, to elucidate the model's predictions. In contrast, \emph{PGExplainer}~\cite{luo2020parameterized} develops a parameterized mask generator that produces approximated discrete masks to interpret the predictions more effectively. \emph{SubgraphX}~\cite{yuan2021subgraphx} delves into subgraph-level interpretation for GNNs, leveraging the Monte Carlo tree search algorithm~\cite{silver2017mastering} to identify the most important subgraph for a trained model with efficient node pruning. 
When adapting these methods to GDL, the perturbations are conducted at point-level instead of edge-level or subgraph-level.

\vspace{1mm}
\noindent\textbf{Surrogate Methods}~\cite{vu2020pgm, huang2020graphlime, zhang2020relex} employ an interpretable surrogate model to locally approximate the
predictions of the complex ML models. 
\emph{PGM-Explainer}~\cite{vu2020pgm} builds a probabilistic graphical
model to fit the local dataset and to interpret the predictions of the original GNN model. 
The adaption of this method is natural due to the fundamental similarity in the structural representation of data across these domains. Specifically, point clouds in GDL are analogously treated as graphs within GNNs.
Other related surrogate methods include \emph{GraphLime}~\cite{huang2020graphlime} and \emph{RelEx}~\cite{zhang2020relex}, which are not incorporated into our study due to their specific design for node classification tasks.



\subsubsection{Self-Interpretable Methods}
Different from post-hoc methods, self-interpretable methods may design new self-interpretable modules and integrate such modules into existing backbone models. The combined models are then trained from scratch and are interpretable. We adapt the self-interpretable methods for GNNs to GDL models. 

\vspace{1mm}
\noindent\textbf{Attention-Based methods}~\cite{ranjan2020asap} utilize the values of attention weights to identify important input patterns. 
For example, \emph{ASAP}~\cite{ranjan2020asap} captures the importance of each node in
a given graph, preserving the hierarchical
graph structure information by iteratively performing score generation, node selection, and graph coarsening. Other typical attention-based methods include \emph{GAT}~\cite{velivckovic2017gat} and \emph{GATv2}~\cite{ma2021gatv2}. We extend these attention weights from nodes in graphs to points in point clouds. 

\vspace{1mm}
\noindent\textbf{IB-Induced Methods}~\cite{yu2022vgib, miao2022lri} usually inject noise to restrict the flow of information and encourage the model to learn to denoise the data, preserving the most relevant information. \emph{VGIB}~\cite{yu2022vgib} injects noise into
the node representations via a learned probability for each node. Instead of perturbing representations, \emph{LRI}-induced methods~\cite{miao2022lri} perturbs inputs by sampling stochastic noise from a learnable distribution, where the distribution can be formulated as a Bernoulli distribution to perturb the existence of input points or as a Gaussian distribution to perturb geometric features. \emph{GIB}~\cite{yu2020graph} is also an IB-induced method that presents as a preliminary work to VGIB and thus has not been included in our benchmark.

\vspace{1mm}
\noindent\textbf{Causality-Based Methods}~\cite{chen2022ciga} assume that the causal relationships within the data remain unchanged across different environments. 
Built upon three structural causal models, \emph{CIGA}~\cite{chen2022ciga} aims to maximize the mutual information of graphs in the dataset with the same label and utilizes contrastive learning with supervised sampling for approximation and optimization. \emph{DIR}~\cite{wu2022discovering} also belongs to the category of causality-based methods.

\subsection{Scientific Applications and Datasets}\label{subsec:datasets}

Since one of our ultimate goals is to promote scientific discovery, we employ datasets that are derived from real-world scientific applications. These datasets not only have a class label for each sample for model training but also provide labels for each point indicating the ground-truth \decisive that determine the class label according to specific scientific principles for evaluating interpretability methods. Below we introduce the 4 datasets used with more details.





\vspace{1mm}
\noindent\textbf{ActsTrack}~\cite{ai2022common} is a dataset for particle tracking in HEP, focusing on the reconstruction of charged particles' properties through position measurements from tracking detectors.  This process is vital for identifying particle types, reconstructing collision events, suppressing background noise, and isolating rare events of interest.
The output of this task serves as the fundamental input of many downstream analyses in HEP experiments~\cite{thomson2013modern, ai2022common}. 
In the context of evaluating interpretability methods, the task is reformulated to predict the occurrence of a $z \rightarrow \mu\mu$ decay within each point cloud $C$. Here, each sample $C$ is associated with a binary class label $Y$ indicating the presence or absence of the $z \rightarrow \mu\mu$ decay, and the points representing the $\mu$'s from the decay are labeled as ground-truth \decisive, since their existence directly indicates the occurrence of the $z \rightarrow \mu\mu$ decay.

\vspace{1mm}
\noindent\textbf{Tau3Mu}~\cite{cms2020search} is 
another HEP dataset aiming at evaluating algorithms designed to detect the rare and challenging signature of charged lepton flavor-violating decays, specifically the $\tau \rightarrow \mu\mu\mu$ decay, using simulated data of muon detector hits from proton-proton collisions.
These decays are highly suppressed in the Standard Model of particle physics~\cite{oerter2006theory, Blackstone_2020}, making their detection indicative of new physics beyond the Standard Model~\cite{calibbi2018charged, atlas2020search}. However, the $\tau \rightarrow \mu\mu\mu$ events are predicted to occur at an extremely low rate, approximately at a branching fraction of $10^{-8}$, rendering it impractical to collect ample real experiment data for training ML models. Consequently, research in this direction leverages carefully calibrated simulation algorithms to generate labeled data for model training.
This reliance underscores the critical importance of validating the trustworthiness of trained models in the context of high-stakes LHC experiments. 
Within the \taumu dataset, each point cloud sample $C$ is assigned a binary class label $Y$, indicating whether or not there occurs a $\tau \rightarrow \mu\mu\mu$ event in $C$. To evaluate interpretability methods, this dataset has also labeled the points in each $C$ that represent the $\mu$'s from the decay as ground-truth \decisive.

\vspace{1mm}
\noindent\textbf{SynMol}~\cite{mccloskey2019using} centers on molecular property prediction, a critical task to accelerate the discovery and development of new materials and drugs. By learning from vast datasets of molecules and their properties in a data-driven way, ML algorithms may predict the properties of unseen molecules with high accuracy, which can surpass traditional computational methods in both speed and precision~\cite{faber2017prediction}. Effective ML models for this task enable scientists to efficiently screen countless compounds and identify promising candidates for in-depth analysis, significantly reducing the time and cost needed by experimental testing. 
Nevertheless, the opaque nature of many ML models presents challenges in understanding the underlying reasons for the predicted properties, and interpretable methods for this task extend the objective beyond merely accurate property prediction of individual molecules but also require the identification of critical data patterns (e.g., certain functional groups) that induce the predicted properties, which can further enrich our understanding and guide future discoveries. For the \syn dataset, the task is to predict molecules' properties determined by two functional groups: carbonyl and unbranched alkane~\cite{mccloskey2019using}, and atoms within these functional groups are labeled as \decisive.

\vspace{1mm}
\noindent\textbf{PLBind}~\cite{wang2005pdbbind} focuses on predicting protein-ligand binding affinities. This task is crucial for drug discovery and design, as understanding how well a drug (ligand) binds to its target protein can inform the efficacy of a potential therapeutic. ML models that can predict binding affinities accurately can significantly facilitate the drug development pipeline by replacing the less efficient traditional docking simulations.
Interpretable ML models are invaluable for elucidating the complex mechanisms of protein-ligand interactions, specifically by identifying the critical regions of interaction, or binding sites, on the protein surface. This insight is instrumental in understanding the binding mechanism, guiding the rational design of more effective and targeted therapeutics. For the \plb dataset, binary classifiers are trained to predict protein-ligand pairs with high or low binding affinities, and those amino acids near the binding site are labeled as ground-truth \decisive for evaluating interpretability methods.

\begin{table*}[!tbp]
\centering
\caption{Definition of metrics, where $n$ is the total number of samples in a dataset, $y_i$ denotes the ground-truth class label of the $i^{th}$ sample $C^{(i)}$, $\hat{y}_i$ denotes the predicted label yielded using the raw input $C^{(i)}$, $\hat{y}_i^{\rho+}$ is the predicted label yielded using $C^{(i)}\backslash C_s^{(i)}$ and the size of $C_s^{(i)}$ is determined by $\rho$, and $\hat{y}_i^{\rho-}$ is the prediction output yielded using $C_s^{(i)}$ given a specified $\rho$. Additionally, $\mathbbm{1}(\cdot)$ is an indicator function that outputs $1$ if the specified condition is true, and $0$ otherwise; $\operatorname{AUC}(\cdot)$ computes the area under a given curve.
}
\label{tab:metrics}
\renewcommand\arraystretch{1.2}
\resizebox{\linewidth}{!}
{\begin{tabular}{lcl}
\toprule
Metric Name & Definition & Measurement\\
\midrule
Fidelity+@$\rho$ & 
    $\frac{1}{n}\sum_{i}^{n}(\mathbbm{1}(\hat{y}_i = y_i) - \mathbbm{1}(\hat{y}_{i}^{\rho+} = y_i))$ &  The impact of $C\backslash C_s$ \\
Fidelity-@$\rho$ & 
    $\frac{1}{n}\sum_{i}^{n}(\mathbbm{1}(\hat{y}_i  y_i) - \mathbbm{1}(\hat{y}_{i}^{\rho-} = y_i))$ & The impact of $C_s$\\
\fidelityp & $\operatorname{AUC}(\text{Fidelity+ Curve})$ & The averaged impact of $C\backslash C_s$ across various sizes of $C_s$\\
\fidelityn & $\operatorname{AUC}(\text{Fidelity- Curve})$ &  The averaged impact of $C_s$ across various sizes\\
\fidelity & $(\text{\fidelityp} - \text{\fidelityn} + 1)/2$ & The overall impact of the identified important points\\
\bottomrule
    \end{tabular}}
\end{table*}


\subsection{More Specifics on Experiment Settings}\label{sec:exp}
\textbf{Evaluation Metrics.} 
To evaluate when the model interpretations are aligned with \sensitive, based on the importance score assigned to each point, a subset of critical points $C_s$ is identified by selecting the top-ranked points, and the number of points to be selected is determined by the selection ratio $\rho$, e.g., for $\rho=0.2$, the top-ranked 20\% critical points in $C$ will form $C_s$. Then, we compute Fidelity+@$\rho$ and Fidelity-@$\rho$~\cite{yuan2021subgraphx, yuan2022explainability} to measure the impact of $C_s$ on the model's predictive behavior by collecting the changes in prediction outputs when only unimportant part $C\backslash C_s$ (Fidelity+) or important part $C_s$ (Fidelity-) is included as inputs. For a more holistic evaluation of the impact of $C_s$ with different sizes, we compute Fidelity+@$\rho$ and Fidelity-@$\rho$ at different $\rho \in \{0.2, 0.3, 0.4, 0.5, 0.6, 0.7, 0.8, 0.9 \}$, each resulting in a curve with the x-axis being the value of $\rho$ and y-axis being the corresponding Fidelity+@$\rho$ or Fidelity-@$\rho$. Thus, Fidelity+ AUC and Fidelity- AUC are computed according to the area under each curve. Finally, Fidelity AUC is yielded by combining both Fidelity+ AUC and Fidelity- AUC. The formal definition of these metrics is summarized in Table~\ref{tab:metrics}.
To evaluate the identification of \decisive, we directly compare the obtained importance score for each point (i.e., $\mathcal{W}$) and the labeled ground-truth \decisive (i.e., $\mathcal{I}$) and compute ROC-AUC, reported as Interpretation ROC-AUC.

\section*{Data and software availability}
Datasets used in this study are freely available on Zenodo at \url{https://doi.org/10.5281/zenodo.7265547}. The source code of this study is publicly available on Github at \url{https://github.com/Graph-COM/xgdl}.

\section*{Acknowledgements}
This work is supported by the National Science Foundation (NSF) awards PHY-2117997 and IIS-2239565. 
\bibliographystyle{ieeetr}
\bibliography{new_ref}

\pagebreak
\setcounter{figure}{0} 
\setcounter{equation}{0} 
\setcounter{table}{0} 
\setcounter{page}{1} 
\setcounter{section}{0}




\captionsetup[table]{labelfont={bf},labelformat={default},labelsep=period,name={Supplementary Table},list=yes}

\section{Classification Performance of the Trained Models}
Most experiments in the main manuscript use 10 models trained with different seeds via empirical risk minimization (ERM) or objectives proposed by self-interpretable methods. As suggested by our findings, prediction performance may influence interpretation performance. Therefore, below we report these models' prediction performance.

\begin{table*}[h]%
\centering
\renewcommand{\arraystretch}{0.9}
\resizebox{\linewidth}{!}{\begin{tabular}{@{}lcccccc@{}}%
\toprule%
\multicolumn{1}{c}{\multirow{2}{*}{\syn}}&\multicolumn{3}{c}{Classification Accuracy}&\multicolumn{3}{c}{Classification ROC-AUC}\\%
\cmidrule(lr){2-4} \cmidrule(lr){5-7}%
\multicolumn{1}{c}{}&EGNN&DGCNN&PointTrans&EGNN&DGCNN&PointTrans\\%
\midrule%
ERM&$99.45\pm0.16$&$99.40\pm0.25$&$94.63\pm0.54$&$99.87\pm0.08$&$99.99\pm0.01$&$97.96\pm0.40$\\%
ASAP&$95.16\pm8.04$&$98.66\pm0.44$&$87.45\pm5.32$&$93.07\pm14.26$&$99.77\pm0.09$&$80.67\pm20.86$\\%
CIGA&$89.27\pm3.87$&$84.41\pm2.02$&$83.49\pm1.97$&$92.85\pm3.93$&$84.55\pm3.55$&$83.44\pm1.62$\\%
LRI-Bern&$98.72\pm0.47$&$98.91\pm0.30$&$93.29\pm0.81$&$99.76\pm0.07$&$99.87\pm0.04$&$96.91\pm0.82$\\%
LRI-Gaussian&$99.08\pm0.23$&$99.19\pm0.16$&$93.68\pm0.61$&$99.87\pm0.03$&$99.93\pm0.03$&$97.36\pm0.52$\\%
VGIB&$95.78\pm5.06$&$91.11\pm8.65$&$91.65\pm1.22$&$99.75\pm0.08$&$99.55\pm0.19$&$96.10\pm0.86$\\\bottomrule%
\end{tabular}}%
\vspace{1mm}

\resizebox{\linewidth}{!}{\begin{tabular}{@{}lcccccc@{}}%
\toprule%
\multicolumn{1}{c}{\multirow{2}{*}{\acts}}&\multicolumn{3}{c}{Classification Accuracy}&\multicolumn{3}{c}{Classification ROC-AUC}\\%
\cmidrule(lr){2-4} \cmidrule(lr){5-7}%
\multicolumn{1}{c}{}&EGNN&DGCNN&PointTrans&EGNN&DGCNN&PointTrans\\%
\midrule%
ERM&$94.85\pm0.71$&$94.93\pm1.25$&$92.22\pm1.30$&$98.74\pm0.37$&$99.30\pm0.16$&$98.06\pm0.41$\\%
ASAP&$93.14\pm1.70$&$93.39\pm2.10$&$92.79\pm1.62$&$98.07\pm0.99$&$98.90\pm0.47$&$97.78\pm0.57$\\%
CIGA&$93.45\pm2.19$&$84.70\pm6.32$&$90.35\pm8.13$&$98.32\pm0.66$&$95.70\pm2.37$&$97.79\pm0.80$\\%
LRI-Bern&$95.09\pm0.76$&$94.46\pm1.20$&$93.18\pm1.32$&$98.72\pm0.46$&$98.97\pm0.26$&$98.31\pm0.28$\\%
LRI-Gaussian&$95.67\pm0.84$&$94.70\pm0.88$&$94.19\pm2.04$&$99.39\pm0.15$&$99.07\pm0.27$&$98.88\pm0.49$\\%
VGIB&$49.14\pm19.26$&$74.17\pm14.35$&$90.66\pm4.05$&$96.70\pm1.01$&$98.65\pm0.31$&$97.63\pm0.92$\\\bottomrule%
\end{tabular}}%
\vspace{1mm}

\resizebox{\linewidth}{!}{\begin{tabular}{@{}lcccccc@{}}%
\toprule%
\multicolumn{1}{c}{\multirow{2}{*}{\taumu}}&\multicolumn{3}{c}{Classification Accuracy}&\multicolumn{3}{c}{Classification ROC-AUC}\\%
\cmidrule(lr){2-4} \cmidrule(lr){5-7}%
\multicolumn{1}{c}{}&EGNN&DGCNN&PointTrans&EGNN&DGCNN&PointTrans\\%
\midrule%
ERM&$82.52\pm0.67$&$83.63\pm0.18$&$82.54\pm0.21$&$86.18\pm1.04$&$87.45\pm0.12$&$86.10\pm0.20$\\%
ASAP&$81.45\pm1.03$&$79.15\pm2.76$&$76.72\pm0.83$&$84.27\pm1.46$&$82.76\pm2.79$&$78.27\pm1.02$\\%
CIGA&$80.20\pm0.99$&$81.77\pm1.63$&$81.74\pm0.53$&$85.21\pm0.30$&$85.50\pm0.47$&$85.49\pm0.35$\\%
LRI-Bern&$82.77\pm0.13$&$83.47\pm0.28$&$82.72\pm0.25$&$86.41\pm0.11$&$87.30\pm0.10$&$86.49\pm0.10$\\%
LRI-Gaussian&$83.07\pm0.16$&$83.88\pm0.20$&$82.60\pm0.29$&$86.72\pm0.07$&$87.64\pm0.17$&$85.88\pm0.34$\\%
VGIB&$82.12\pm0.54$&$82.94\pm0.57$&$81.52\pm0.59$&$86.11\pm0.17$&$87.04\pm0.14$&$84.59\pm0.61$\\\bottomrule%
\end{tabular}}%
\vspace{1mm}
\resizebox{\linewidth}{!}{\begin{tabular}{@{}lcccccc@{}}%
\toprule%
\multicolumn{1}{c}{\multirow{2}{*}{\plb}}&\multicolumn{3}{c}{Classification Accuracy}&\multicolumn{3}{c}{Classification ROC-AUC}\\%
\cmidrule(lr){2-4} \cmidrule(lr){5-7}%
\multicolumn{1}{c}{}&EGNN&DGCNN&PointTrans&EGNN&DGCNN&PointTrans\\%
\midrule%
ERM&$85.59\pm1.66$&$82.65\pm2.87$&$81.76\pm3.78$&$88.05\pm1.44$&$82.65\pm5.46$&$81.57\pm2.47$\\%
ASAP&$84.78\pm1.14$&$84.16\pm2.00$&$82.98\pm2.67$&$84.19\pm1.75$&$82.23\pm1.89$&$78.89\pm8.07$\\%
CIGA&$83.71\pm2.39$&$82.33\pm2.36$&$84.37\pm1.35$&$82.60\pm1.48$&$82.51\pm2.12$&$82.37\pm2.10$\\%
LRI-Bern&$85.10\pm1.85$&$84.65\pm2.23$&$82.57\pm2.04$&$85.49\pm3.77$&$84.70\pm2.50$&$85.22\pm1.99$\\%
LRI-Gaussian&$85.80\pm1.33$&$83.35\pm3.32$&$84.29\pm1.77$&$89.17\pm1.40$&$84.43\pm3.93$&$86.26\pm1.81$\\%
VGIB&$82.29\pm2.14$&$77.51\pm8.65$&$78.24\pm5.39$&$78.44\pm4.24$&$78.72\pm2.56$&$79.18\pm3.26$\\\bottomrule%
\end{tabular}}%

\caption{Prediction performance of trained models.}
\label{tab:clf_perf}

\end{table*}

\pagebreak
\section{Misalignment of the Two Patterns}
In the main manuscript, we discussed that the relatively low and highly variable \fidelity may indicate a misalignment between the two patterns. Moreover, the \labelfidelity is significantly lower than the highest \fidelity achieved by post-hoc methods, thereby reinforcing our claim.

\begin{table*}[h]%
\centering%

\resizebox{\linewidth}{!}{
\begin{tabular}{@{}lcccccc@{}}
\toprule
\multicolumn{1}{c}{\multirow{2}{*}{Dataset}} & \multicolumn{3}{c}{Best Fidelity AUC} & \multicolumn{3}{c}{\labelfidelity}  \\
\cmidrule(lr){2-4} \cmidrule(lr){5-7}
{} & EGNN & DGCNN & PointTrans & EGNN & DGCNN & PointTrans \\
\midrule
\syn & $77.77$ &  $24.27$ & $63.94$ & $61.91 \pm 10.95$ & $1.75 \pm 3.38$ & $17.62 \pm 3.79$ \\
\acts & $90.12$ & $84.76$ & $93.69$ & $85.46 \pm 2.48$ & $58.48 \pm 6.05$ & $70.97 \pm 2.81$ \\
\taumu & $24.71$ & $27.43$ & $25.94$ & $17.04 \pm 2.89$ & $21.81 \pm 1.93$ & $20.07 \pm 1.18$ \\
\plb & $72.86$ & $53.66$ & $57.29$ & $20.71 \pm 6.68$ & $20.31 \pm 4.66$ & $8.46 \pm 5.71$ \\
\bottomrule
\end{tabular}}

\caption{The degree of alignment between the two patterns measured by \labelfidelity. The best \fidelity is from the best-performing post-hoc methods benchmarked in the main manuscript.
}
\label{tab:notalign}
\end{table*}

\pagebreak
\section{Ensemble Strategy Improves the Alignment}
In the main manuscript, we demonstrated that our ensemble strategy, which combines multiple pre-trained models, is effective in enhancing the alignment between post-hoc interpretations and the \decisive. To differentiate our approach from the naive ensemble methods commonly utilized in machine learning, we conducted a comparative study, involving a naive ensemble applied to multiple explainers yielded with different seeds but for a single trained model (instead of different trained models). The comparative analysis substantiates our assertion that the success of our ensemble strategy stems from the overlapping of \sensitive across various already-trained models, which yield interpretations more aligned with \decisive.

\begin{table*}[h]
\centering

\resizebox{\linewidth}{!}{\begin{tabular}{@{}lcccccc@{}}%
\toprule%
\multicolumn{1}{c}{\multirow{2}{*}{SynMol}}&\multicolumn{3}{c}{Ensemble Model}&\multicolumn{3}{c}{Ensemble Explainer}\\%
\cmidrule(lr){2-4} \cmidrule(lr){5-7}%
\multicolumn{1}{c}{}&EGNN&DGCNN&PointTrans&EGNN&DGCNN&PointTrans\\%
\midrule%
GNNExplainer&$70.98$&$45.49$&$39.92$&$58.77\pm16.94$&$51.14\pm6.37$&$26.86\pm2.93$\\%
PGExplainer&$96.20$&$94.69$&$90.51$&$81.36\pm22.98$&$42.49\pm39.38$&$89.14\pm1.11$\\%
PGM-Explainer&$68.83$&$53.31$&$63.94$&$67.84\pm1.53$&$51.55\pm2.33$&$62.06\pm1.42$\\%
SubgraphX&$92.32$&$76.84$&$82.38$&$90.29\pm1.30$&$71.12\pm8.35$&$79.11\pm1.26$\\\bottomrule%
\end{tabular}}%
\vspace{2mm}
\resizebox{\linewidth}{!}{\begin{tabular}{@{}lcccccc@{}}%
\toprule%
\multicolumn{1}{c}{\multirow{2}{*}{ActsTrack}}&\multicolumn{3}{c}{Ensemble Model}&\multicolumn{3}{c}{Ensemble Explainer}\\%
\cmidrule(lr){2-4} \cmidrule(lr){5-7}%
\multicolumn{1}{c}{}&EGNN&DGCNN&PointTrans&EGNN&DGCNN&PointTrans\\%
\midrule%
GNNExplainer&$62.47$&$75.73$&$80.08$&$52.03\pm4.59$&$64.43\pm4.07$&$71.57\pm2.66$\\%
PGExplainer&$63.59$&$95.01$&$91.71$&$34.95\pm26.61$&$92.91\pm1.65$&$89.19\pm2.17$\\%
PGM-Explainer&$70.26$&$58.52$&$61.58$&$68.28\pm1.25$&$56.84\pm2.14$&$59.13\pm2.13$\\%
SubgraphX&$64.60$&$60.21$&$63.93$&$62.64\pm2.28$&$59.82\pm1.20$&$63.02\pm1.06$\\\bottomrule%
\end{tabular}}%
\vspace{2mm}
\resizebox{\linewidth}{!}{\begin{tabular}{@{}lcccccc@{}}%
\toprule%
\multicolumn{1}{c}{\multirow{2}{*}{Tau3Mu}}&\multicolumn{3}{c}{Ensemble Model}&\multicolumn{3}{c}{Ensemble Explainer}\\%
\cmidrule(lr){2-4} \cmidrule(lr){5-7}%
\multicolumn{1}{c}{}&EGNN&DGCNN&PointTrans&EGNN&DGCNN&PointTrans\\%
\midrule%
GNNExplainer&$77.52$&$57.22$&$41.64$&$72.17\pm2.44$&$52.36\pm3.33$&$31.14\pm0.80$\\%
PGExplainer&$78.60$&$71.90$&$80.01$&$76.67\pm1.57$&$51.23\pm20.35$&$79.28\pm0.29$\\\bottomrule%
\end{tabular}}%
\caption{Interpretation ROC-AUC performance of different ensemble schemes for post-hoc methods: Ensemble Model refers to the setting reported in our main manuscript, where the explainers are trained with the same seed as the models. Ensemble Explainer refers to a setting where the already-trained model is fixed (trained with seed 0), and we ensemble multiple explainers trained with different seeds. The results are reported as mean $\pm$ standard deviation.}
\label{tab:more_ens}
\end{table*}

\pagebreak

\section{Post-Hoc Interpretations on Self-Interpretable Models}
In the main manuscript, we demonstrated that post-hoc methods can produce interpretations that align better with \decisive when applied to self-interpretable models rather than vanilla models trained with ERM, indicating that \sensitive of self-interpretable models may align well with \decisive. For simplicity, we only showed the average results across three backbones, and below are the complete results.

\begin{table*}[h]%
\centering

\resizebox{\linewidth}{!}{\begin{tabular}{@{}lccccccc@{}}%
\toprule%
\multicolumn{2}{c}{\multirow{2}{*}{Model \& Explainer}}&\multicolumn{3}{c}{\syn}&\multicolumn{3}{c}{\acts}\\%
\cmidrule(lr){3-5} \cmidrule(lr){6-8}
\multicolumn{2}{c}{}&EGNN &DGCNN&Point Transformer&EGNN &DGCNN&Point Transformer\\%
\midrule%
\multirow{4}{*}{Pre-trained Models}&GNNLRP&$81.75\pm4.01$&$84.61\pm3.81$&$50.38\pm1.68$&$86.01\pm2.31$&$86.40\pm5.08$&$50.2\pm1.93$\\%
\multirow{4}{*}{}&GradCAM&$57.82\pm4.42$&$78.89\pm3.84$&$84.1\pm3.66$&$69.38\pm2.72$&$75.25\pm3.67$&$77.32\pm2.83$\\%
\multirow{4}{*}{}&GradxInput&$76.03\pm4.82$&$71.39\pm5.89$&$78.03\pm1.52$&$68.74\pm1.84$&$65.17\pm1.56$&$64.78\pm1.9$\\%
\multirow{4}{*}{}&IG&$78.59\pm7.83$&$64.31\pm9.14$&$84.23\pm1.83$&$68.78\pm1.82$&$65.27\pm1.47$&$64.8\pm1.88$\\%
\multirow{4}{*}{}&GNNExplainer&$58.94\pm15.89$ & $51.03\pm5.58$ & $26.28\pm2.70$&$51.77\pm4.41$&$64.34\pm4.05$&$71.38\pm2.61$\\%
\multirow{4}{*}{}&PGExplainer&$77.92\pm22.04$&$49.56\pm39.82$&$87.41\pm2.66$&$33.54\pm23.17$&$92.63\pm1.57$&$88.39\pm3.13$\\%

\midrule

\multirow{7}{*}{LRI-Bern Induced Models}&GNNLRP&$80.96\pm3.88$&$84.33\pm4.28$&$50.39\pm1.37$&$88.06\pm4.24$&$89.16\pm3.04$&$50.21\pm1.13$\\%
&GradCAM&$57.03\pm8.16$&$78.08\pm4.73$&$85.32\pm1.52$&$68.94\pm3.63$&$80.7\pm2.32$&$82.33\pm2.87$\\%
&GradxInput&$87.85\pm6.86$&$82.49\pm6.16$&$78.87\pm1.06$&$70.56\pm2.15$&$63.71\pm1.38$&$65.3\pm1.66$\\%
&IG&$75.27\pm9.96$&$86.83\pm4.64$&$84.21\pm2.45$&$70.68\pm2.15$&$63.82\pm1.41$&$65.32\pm1.66$\\%
&GNNExplainer&$68.92 \pm0.15$&$46.66\pm0.11$&$25.90 \pm 0.03$&$32.56 \pm 0.04$&$77.55 \pm 0.04$&$80.13 \pm 0.02$\\%
&PGExplainer&$78.89 \pm 0.38$&$74.85 \pm 0.33$&$79.35 \pm 0.25$&$78.60 \pm 0.12$&$\underline{93.51} \pm 0.01$&$\underline{90.16} \pm 0.01$\\%
&Self&$92.04\pm3.0$&$94.2\pm4.53$&$90.46\pm1.21$&$80.97\pm2.07$&$90.74\pm1.72$&$86.84\pm1.85$\\%
\midrule

\multirow{7}{*}{LRI-Gaussian Induced Models}&GNNLRP&$85.37\pm1.16$&$86.41\pm1.37$&$50.47\pm1.69$&$90.48\pm3.88$&$88.66\pm4.33$&$50.08\pm0.92$\\%
&GradCAM&$69.01\pm7.36$&$78.27\pm3.47$&$78.18\pm4.79$&$82.3\pm5.35$&$66.98\pm4.46$&$83.2\pm2.43$\\%
&GradxInput&$83.98\pm7.0$&$85.72\pm3.2$&$80.52\pm2.46$&$73.41\pm1.87$&$74.52\pm1.15$&$75.52\pm3.63$\\%
&IG&$88.73\pm2.1$&$87.21\pm5.86$&$82.48\pm1.85$&$73.42\pm1.77$&$74.55\pm1.17$&$75.54\pm3.64$\\%
&GNNExplainer&$49.79 \pm 0.01$&$49.88 \pm 0.01$&$49.84 \pm 0.01$&$49.82 \pm 0.01$&$50.05 \pm 0.01$&$50.05 \pm 0.01$\\%
&PGExplainer&$36.53 \pm 0.17$&$51.87 \pm 0.12$&$37.97 \pm 0.16$&$83.81 \pm 0.07$&$87.47 \pm 0.04$&$85.50 \pm 0.07$\\%
&Self&$97.13\pm0.79$&$98.23\pm1.0$&$93.06\pm1.19$&$92.93\pm1.58$&$94.18\pm0.88$&$91.85\pm1.15$\\%
\bottomrule%
\end{tabular}}%
\caption{Interpretation ROC-AUC computed using the interpretations given by post-hoc methods for self-interpretable models trained by LRI-induced methods.}
\label{tab:more_post_inherent}
\vspace{2mm}
\end{table*}

\end{document}